\documentclass[10pt,journal,compsoc]{IEEEtran}
\usepackage{amsmath,amsfonts}
\usepackage{algorithmic}
\usepackage{array}
\usepackage{textcomp}
\usepackage{stfloats}
\usepackage{url}
\usepackage{verbatim}
\usepackage{graphicx}
\usepackage[table]{xcolor}
\usepackage{epsfig}
\usepackage{graphicx}
\usepackage{amsmath}
\usepackage{amssymb}
\usepackage{enumitem}
\usepackage{pifont}
\usepackage[linesnumbered,titlenumbered,ruled,vlined,commentsnumbered]{algorithm2e}
\SetKwComment{Comment}{$\triangleright$\ }{}

\usepackage{multirow}
\usepackage{tabularx}
\usepackage{tcolorbox}
\usepackage{dsfont}
\usepackage{url}
\usepackage[tight]{subfigure}
\usepackage{color}
\usepackage{subfigure}
\usepackage{amsthm}
\usepackage[export]{adjustbox}
\usepackage{comment}
\usepackage[utf8]{inputenc} 
\usepackage[T1]{fontenc}    
\usepackage{url}            
\usepackage{booktabs}       
\usepackage{amsfonts}       
\usepackage{nicefrac}       
\usepackage{microtype}      
\usepackage{diagbox}
\usepackage{balance}
\usepackage[colorlinks,linkcolor=red,citecolor=blue]{hyperref}

\usepackage{mathtools}
\usepackage{soul}
\usepackage{stfloats}
\usepackage{colortbl}
\usepackage{bm}
\usepackage{makecell}
\usepackage{subcaption}
\usepackage{sidecap}
\usepackage{helvet}  
\usepackage{courier}  
\usepackage{babel}
\usepackage{array}
\usepackage[capitalize,noabbrev]{cleveref}

\usepackage{xspace}
\makeatletter
\DeclareRobustCommand\onedot{\futurelet\@let@token\@onedot}
\def\@onedot{\ifx\@let@token.\else.\null\fi\xspace}
\def\eg{\emph{e.g}\onedot} 
\def\ie{\emph{i.e}\onedot} 
 
\def\etc{\emph{etc}\onedot}

\makeatother

\ifCLASSOPTIONcompsoc
  \usepackage[nocompress]{cite}
\else
  \usepackage{cite}
\fi
\ifCLASSINFOpdf
\else
\fi
\begin{document}
%
\title{Improving Robustness of LiDAR-Camera \\ Fusion Model against Weather Corruption from Fusion Strategy Perspective}
%
%
%
%

\author{Yihao~Huang,
        Kaiyuan~Yu,
        Qing~Guo,
        Felix~Juefei-Xu,
        Xiaojun~Jia,
        Tianlin~Li,
        Geguang~Pu,
        and~Yang~Liu
}
\author{
    \IEEEauthorblockN{Yihao~Huang$^1$,
        Kaiyuan~Yu$^2$,
        Qing~Guo$^3$,
        Felix~Juefei-Xu$^4$,
        Xiaojun~Jia$^1$,
        Tianlin~Li$^1$,\\
        Geguang~Pu$^2$,
        and~Yang~Liu$^1$}\\
    \IEEEauthorblockA{$^1$ Nanyang Technological University, Singapore}\\
    \IEEEauthorblockA{$^2$ East China Normal University, China}\\
    \IEEEauthorblockA{$^3$ CFAR and IHPC, Agency for Science, Technology and Research (A*STAR), Singapore}\\
    \IEEEauthorblockA{$^4$ New York University, USA}
}

%
%

\markboth{Journal of \LaTeX\ Class Files,~Vol.~14, No.~8, August~2015}%
{Shell \MakeLowercase{\textit{et al.}}: Bare Advanced Demo of IEEEtran.cls for IEEE Computer Society Journals}
%



\IEEEtitleabstractindextext{%
\begin{abstract}
In recent years, LiDAR-camera fusion models have markedly advanced 3D object detection tasks in autonomous driving. However, their robustness against common weather corruption such as fog, rain, snow, and sunlight in the intricate physical world remains underexplored. In this paper, we evaluate the robustness of fusion models from the perspective of fusion strategies on the corrupted dataset.
Based on the evaluation, we further propose a concise yet practical fusion strategy to enhance the robustness of the fusion models, namely flexibly weighted fusing features from LiDAR and camera sources to adapt to varying weather scenarios. Experiments conducted on four types of fusion models, each with two distinct lightweight implementations, confirm the broad applicability and effectiveness of the approach.
\end{abstract}

}

\maketitle

\IEEEdisplaynontitleabstractindextext

%
\IEEEpeerreviewmaketitle

\ifCLASSOPTIONcompsoc
\IEEEraisesectionheading{\section{Introduction}\label{sec:introduction}}\label{sec:intro}
\else
\section{Introduction}
\label{sec:introduction}\label{sec:intro}
\fi

\IEEEPARstart{I}n the field of autonomous driving, 3D object detection stands as a crucial foundational task, aiming to precisely locate objects on the road and furnish essential perceptual information for subsequent decision-making processes. Over recent years, benefiting from the complementary between semantic and depth information provided by the camera and LiDAR respectively, the LiDAR-camera fusion models \cite{virconv,transfusion,VFF,bevfusion,focalsconv} have surpassed the single-modal model \cite{li2022bevformer,fcos3d,lang2019pointpillars,pvrcnn,shi2019pointrcnn}, exhibiting exceptional performance in the realm of object detection. 

Considering its promising potential, evaluating the applicability of LiDAR-camera fusion models under complex road scenes such as weather corruption is necessary since robustness is an important property in the safety-critical autonomous driving domain. Previous studies \cite{benchmarkingcorruptions,understanding_bev_robust} have broadly assessed the robustness of both single-modal and LiDAR-camera fusion models against weather corruption. However, in their work, there lacks a clear rationale for the selection of LiDAR-camera fusion models and the evaluation is coarse-grained. 

In this paper, we conducted a detailed exploration and evaluation of the crucial properties (\ie, fusion strategy) of the LiDAR-camera fusion model that is nonexistent in the single-modal model. To be specific, we evaluate the fusion strategy from two perspectives. (1) Inspired by the contrast \cite{benchmarkingcorruptions} between the robustness of single-modal (\ie, LiDAR-only and camera-only) models against weather corruption, we have a question that \textit{do fusion models with different fusion modals exhibit distinct robustness?} (2) Considering the significance of LiDAR and camera branches as primary sources for fusion, \textit{to what extent does each corrupted branch impact the robustness of the model}?

\begin{figure}[tb]
\centering
\includegraphics[width=1\linewidth]{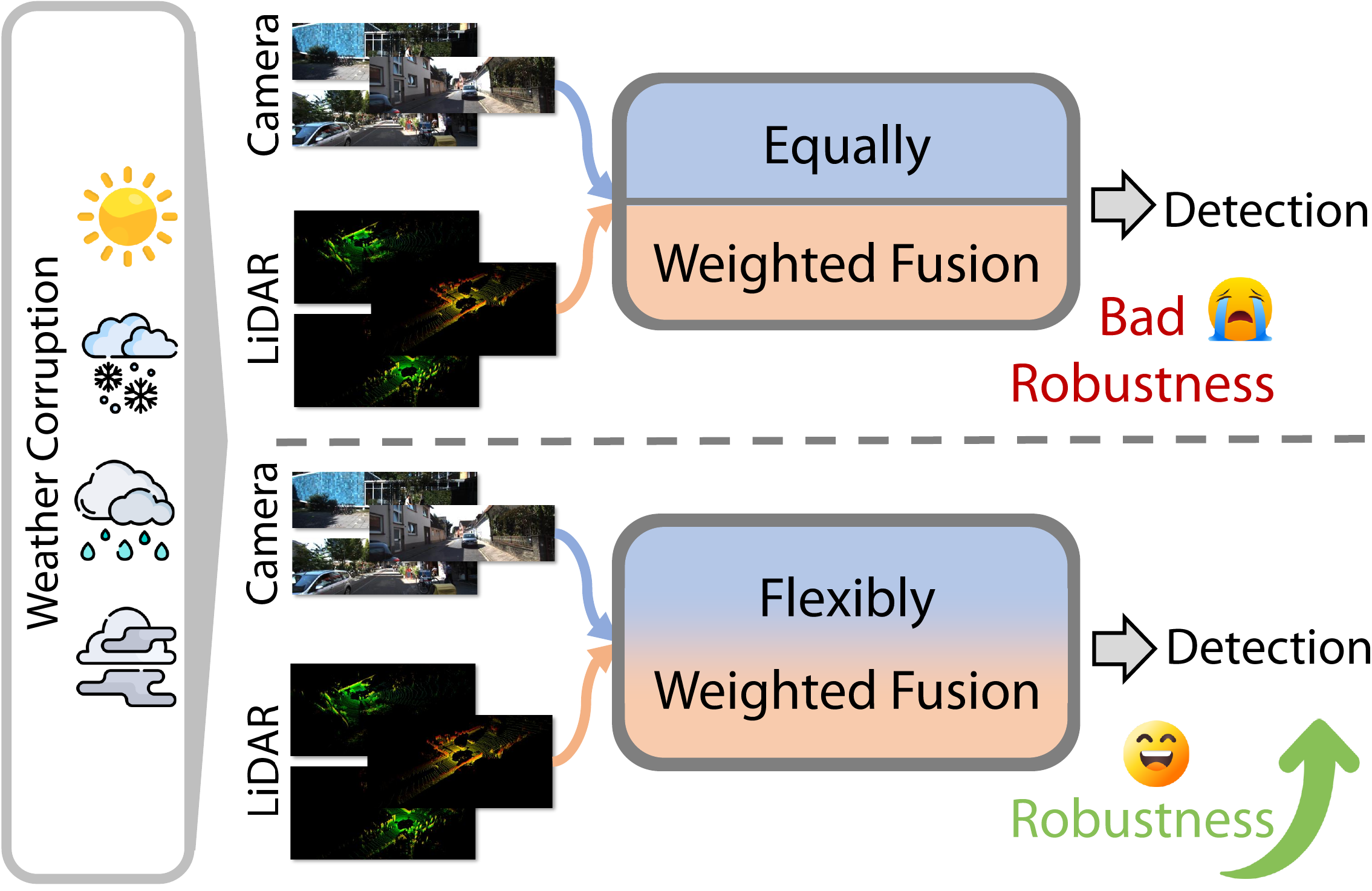}
\caption{Flexible weighted fusion between LiDAR and camera branches of the model improves the robustness of the LiDAR-camera fusion model against weather corruption.}
\label{fig:teaser}
\end{figure}

To answer the first question, we select four classical types of fusion models according to their fusion modal and reveal the virtual point-based fusion model generally shows superior robustness. It is noteworthy that each model demonstrates varying degrees of robustness in response to different weather corruption. For instance, the BEV-based fusion model exhibits good robustness against sunlight but performs poorly in rainy. Additionally, fusion models utilizing different fusion modals display distinct levels of robustness against the same weather corruption. For instance, the virtual point-based fusion model demonstrates strong robustness against fog, whereas the voxel-based fusion model does not.
For the second question, we design experiments by applying weather corruption on only one of the LiDAR and camera branches of the fusion model. Our observations reveal significant differences in the impact on model robustness due to distinct information loss in LiDAR and camera under varying weather corruption. For instance, the corrupted camera branch just minorly influences the robustness of the fusion model against rain and snow, whereas the corrupted LiDAR branch does not. The detailed analysis and conclusion are in Sec.~\ref{sec:evaluation}.

Based on the evaluation and observation, we propose a simple and general yet effective robustness enhancement method against weather corruption for LiDAR-camera fusion models. The high-level idea is flexibly weighted fusing the features from the LiDAR and camera branches instead of simply combining them equally.

To sum up, our work has the following contributions:
\begin{itemize}
\setlength\itemsep{0em}
\item To the best of our knowledge, we are the first to conduct a fine-grained evaluation of the robustness of LiDAR-camera fusion models against weather corruption, specifically focusing on the fusion strategy.
\item Based on the evaluation, we propose a simple yet practical high-level idea to enhance the robustness of the LiDAR-camera fusion models, namely dynamically fusing the features from the LiDAR and camera branches to adapt to different weather corruption.
\item The experiments conducted on four types of fusion modals verify the superior robustness of the virtual point-based fusion model. The two distinct lightweight implementations verify the broad applicability and effectiveness of the proposed enhancement idea.
\end{itemize}

\section{Related Work}
\subsection{LiDAR-camera Fusion Detection}


Since LiDAR and camera features contain complementary information which usually boosts detection performance, multi-modal fusion detection methods (\ie, LiDAR-camera fusion detection) \cite{Multi-modal_servey} have caught the attention of researchers and become popular in the 3D detection domain. AVOD \cite{AVOD}, MV3D \cite{mv3d}, and F-PointNet \cite{F-Pointnet} were early proposals for multimodal fusion. They independently perform feature extraction for two modalities and directly concatenate multimodal features through 2D and 3D RoIs. Point-Painting \cite{pointpainting} and Point-Augmenting \cite{pointaugmenting} represent two techniques for point embellishment, aiming to enhance each LiDAR point using extracted 2D features. Recent efforts have focused on independently encoding features from two modalities and merging them in the same modal or feature space. BEVFusion \cite{bevfusion} and 3D-CVF \cite{3D-CVF} methods encode image information into BEV (Bird's Eye View) maps and fuse them with BEV maps obtained from the original 3D point cloud. MVP \cite{mvp} and VirConv \cite{virconv} achieve fusion by generating virtual points to combine the two modals.


\subsection{Robustness of LiDAR-camera Fusion Methods against Weather Corruption}\label{sec:related_work_weather_robustness}
Robustness, which has been one of the key issues for neural networks, is highly susceptible to common corruption in both 2D \cite{2d_robustness} and 3D domains \cite{robo3d}. Recently, several empirical studies \cite{benchmarkingcorruptions,understanding_bev_robust} have evaluated the robustness against weather corruption of 3D detection methods, including Camera-based, LiDAR-based, and LiDAR-camera fusion models. However, the selection of LiDAR-camera fusion models has no rationale and they only coarsely evaluate the robustness but lack \textit{analysis on the impact of \textbf{fusion strategy} on robustness and how to \textbf{improve the robustness} of LiDAR-camera fusion models}.

\section{Evaluation}\label{sec:evaluation}
\subsection{Weather corruption, Datasets, and Metrics}
\begin{figure}[tb]
\centering
\includegraphics[width=1\linewidth]{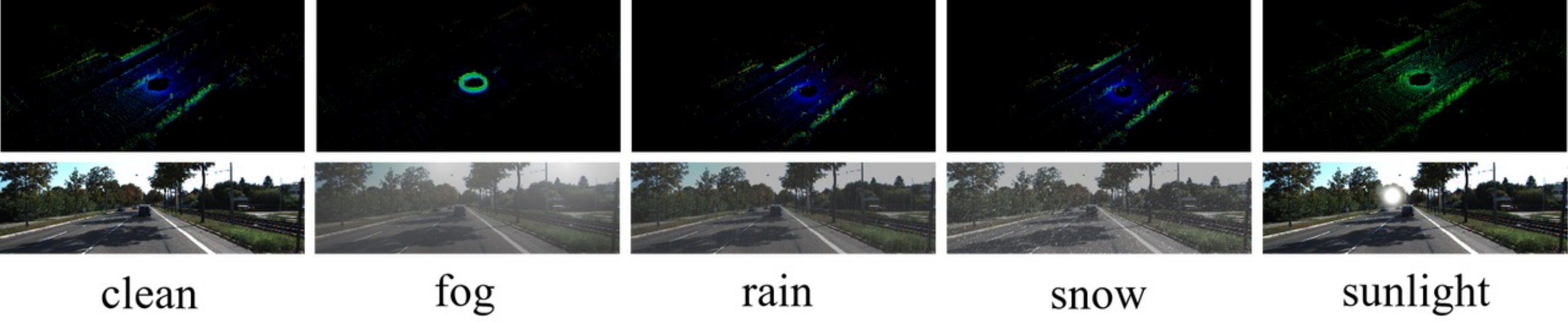}
\caption{Points cloud and image under weather corruption.}
\label{fig:show_corruption}
\end{figure}

\textbf{Weather corruption.} Weather changes often exist in autonomous driving in the physical world, and they have obvious disruptions on information collected by LiDAR and camera \cite{stf}. For example, both the LiDAR and the camera will be simultaneously interfered in rainy weather. Dense droplets will diminish the reflective strength of LiDAR, which leads to signal attenuation and the emergence of scattered points in LiDAR data. Raindrops also cause white lines in the image that interfere with perception \cite{dreissig2023survey,zhang2023perception}. \textbf{In our study, we mainly consider the four most commonly seen weather corruption: snow, rain, fog, and strong sunlight}. For the sake of unification, we use \textbf{four weather corruption} to represent them. To synthesize four weather corruption on point cloud and image, we refer to the physical-based simulation methods used in previous evaluation works \cite{benchmarkingcorruptions} and details are in Section \ref{sec:weather_details}. We visualize corruption on image and point clouds in Fig.~\ref{fig:show_corruption} as examples. 


\noindent\textbf{Datasets.} 
There are many recent real datasets studying LiDAR sensing in severe weather conditions, such as STF \cite{stf}, CADC \cite{cadc}, Ithaca365 \cite{ithaca365}, \etc. However, as mentioned in previous work \cite{benchmarkingcorruptions}, these real datasets are not suitable as the robustness evaluation benchmark against weather corruption due to the following two disadvantages. (1) The real data under severe weather conditions are very scarce. (2) There is a big distribution gap between the scene of real weather corruption datasets and the scene of real large-scale standard clean training datasets (\ie, different cities, vehicles) \cite{common_corruption}. 

Therefore, some work \cite{benchmarkingcorruptions} intended to synthesize different corruption on clean datasets as the robustness evaluation benchmark. In general, they simulate real-world corruption on common large-scale training datasets (\eg, KITTI~\cite{kitti}, NuScenes~\cite{nuscenes}, Waymo~\cite{waymo}) to measure the robustness. Refer to our \textit{\underline{test dataset}}, to fairly compare the robustness of different fusion models under various weather corruption, we also plan to conduct experiments on a \textit{synthesized test dataset} in which we can evenly balance the amount of different corrupted data. To synthesize the test dataset, the original dataset should contain clean data (\ie, images and point clouds without four weather corruption) as much as possible. Among commonly used datasets in 3D detection, as mentioned in previous work \cite{benchmarkingcorruptions}, NuScenes and Waymo both contain weather corruption (\eg, rain) in their data, which may lead to confusion about weather type in the synthesized data if we adding weather corruption on them, thus we choose KITTI as the original clean dataset. The KITTI dataset contains 3,712 training samples and 3,769 validation samples. We simulated four weather corruption on the validation set to construct the synthesized dataset KITTI-C and evaluate the performance of existing fusion models on it. Note that to conduct a more comprehensive exploration of the impact of weather on robustness, we use $low$ and $high$ to represent different corruption severity. The details of weather corruption-related parameters are shown in Section \ref{sec:weather_details}. Refer to our \textit{\underline{training dataset}}, since we synthesize the test dataset with KITTI, thus the training dataset of all the fusion models are the 3,712 samples in the training dataset of KITTI.

\noindent\textbf{Metrics.} 
We used a widely used evaluation metric: average precision (AP) at 40 recall thresholds (R40). We use 3D AP which focuses on the performance of the 3D bounding box, and the BEV AP which focuses on the performance of the rectangular bounding box in the bird's eye view. The combined evaluation of these two metrics provides a comprehensive understanding of target detection algorithms under different detection schemes. In general, the evaluation criteria on AP for the different levels of difficulty of the target detection tasks are categorized as easy, medium, and difficult. The experiment results on medium criteria are shown in the paper while the others are in Section \ref{sec:more_results}. We draw figures to show the relative corruption error (RCE) that can directly reflect the performance reduction of the fusion model between clean and corrupted datasets. The formula is $RCE=\frac{AP - AP_c}{AP}$, where AP and AP$_{c}$ are the performance of the model on the clean and corrupted dataset respectively.

\subsection{Type of LiDAR-camera Fusion Models}
Motivated by the fact that different modals (\eg, image and LiDAR) show distinct robustness against weather corruption \cite{benchmarkingfusionmodel,benchmarkingcorruptions}, we have a question that \textit{do fusion models with different fusion modals exhibit distinct robustness?} From a macro perspective, there are only three multi-modal fusion strategies: \ding{182} point cloud-led pipeline (\ie, adding image information to point clouds), \ding{183} image-led pipeline (\ie, adding point cloud information to images), \ding{184} intermediate modal-led pipeline (\ie, transform both images and point clouds to an intermediate modal and analyze on the intermediate modal). Please note that to be more clear, we think in point cloud-led and image-led pipelines, point cloud and image are fusion modals but not intermediate modals since intermediate modals should be modals other than them. For the intermediate modal-led pipeline, the intermediate modals are also fusion modals.

For the point cloud-led pipeline, we choose the SOTA virtual point-based model VirConv (Virtual-SOTA) \cite{virconv} as the target model. For the image-led pipeline, since depth information will lost when transforming point cloud to image, to our best knowledge, there is no work based on the image-led pipeline. For the intermediate modal-led pipeline, we choose the SOTA FocalsConv (Voxel-SOTA) \cite{focalsconv}, VFF (Ray-SOTA) \cite{VFF}, 3D-CVF (BEV-SOTA) \cite{3D-CVF} which based on three classical and popular intermediate modal (\ie, voxel modal, ray modal, BEV modal) respectively. 

For these four SOTA LiDAR-camera fusion detection models, we evaluate them on KITTI-C to obtain a preliminary conclusion on which one has the best robustness. Furthermore, since the four SOTA models have different backbones, to make a fair comparison, we set their backbone to a structurally general and concise one (\ie, the backbone of VoxelRCNN \cite{voxelrcnn}) to achieve four new models, named as Virtual-fair, Voxel-fair, Ray-fair, BEV-fair. We also evaluate them on KITTI-C for a more fair evaluation of their robustness against weather corruption.
Please note that since FocalsConv (Voxel-SOTA) is based on the backbone of VoxelRCNN, thus the detection results of Voxel-fair should be consistent with that of FocalsConv. 

To sum up, we evaluate totally eight models (Virtual-SOTA, Voxel-SOTA, Ray-SOTA, BEV-SOTA, Virtual-fair, Voxel-fair, Ray-fair, BEV-fair) on KITTI-C and Voxel-SOTA is the same as Voxel-fair.

\begin{table*}[tb]
\centering
\caption{The evaluation results of eight different 3D object detectors on car category of KITTI-C at medium difficulty.}
\resizebox{1\linewidth}{!}{
\begin{tabular}{l|c|cccccccc|c}
\toprule 
model & clean & fog$_{low}$ & fog$_{high}$ & rain$_{low}$ & rain$_{high}$ & snow$_{low}$ & snow$_{high}$ & sunlight$_{low}$ & sunlight$_{high}$ & average \\ \midrule 
 & \multicolumn{10}{c}{Car 3D AP (R40)}\\ \midrule
Virtual-SOTA (VirConv) & \textbf{87.92} & \textbf{79.92} & \textbf{76.89} & \textbf{61.66} & \textbf{57.02} & \textbf{58.97} & \textbf{55.19} & \textbf{85.86} & \textbf{78.42} & \textbf{69.24}\\ 
Voxel-SOTA (FocalsConv) & 85.52 & 57.52 & 46.71 & 46.29 & 37.74 & 35.95 & 27.13 & 84.79 & 76.01 & 51.52\\ 
Ray-SOTA (VFF) & 85.26 & 74.85 & 73.14 & 54.15 & 49.11 & 56.09 & 46.57 & 81.66 & 62.44 & 62.25\\ 
BEV-SOTA (3D-CVF) & 78.92 & 72.69 & 69.43 & 52.42 & 48.14 & 52.28 & 47.43 & 75.22 & 68.41 & 60.73 \\ \midrule
Virtual-fair  & \textbf{85.78} & 77.76 & 74.55 & \textbf{57.39} & \textbf{51.77} & 50.89 & 46.69 & \textbf{85.49} & \textbf{79.43} & \textbf{65.50}\\
Voxel-fair  & 85.52 & 57.52 & 46.71 & 46.29 & 37.74 & 35.95 & 27.13 & 84.79 & 76.01 & 51.52 \\
Ray-fair  & 85.10 & 77.73 & 75.19 & 45.15 & 44.93 & 46.68 & 45.07 & 82.50 & 72.62 & 61.23 \\
BEV-fair  & 82.45 & \textbf{78.43} & \textbf{75.82} & 53.74 & 51.05 & \textbf{54.96} & \textbf{53.48} & 81.84 & 69.22 & 64.82\\
\midrule 
 & \multicolumn{10}{c}{Car BEV AP (R40)}\\
\midrule 
Virtual-SOTA (VirConv) & \textbf{91.70} & \textbf{88.29} & \textbf{86.27} & \textbf{70.86} & \textbf{64.87} & 67.89 & \textbf{61.75} & \textbf{91.62} & \textbf{85.24} & \textbf{77.10}\\
Voxel-SOTA (FocalsConv) & 91.32 & 66.19 & 52.57 & 55.48 & 42.22 & 42.03 & 30.88 & 88.98 & 82.41 & 57.60\\
Ray-SOTA (VFF) & 88.23 & 85.36 & 82.81 & 66.71 & 62.49 & \textbf{68.81} & 59.33 & 87.97 & 73.31 & 73.35 \\
BEV-SOTA (3D-CVF) & 88.20 & 82.27 & 77.82 & 61.53 & 56.56 & 60.85 & 54.05 & 81.51 & 76.52 & 68.89 \\
\midrule 
Virtual-fair  & \textbf{91.51} & 86.48 & 81.45 & \textbf{66.72} & 55.23 & 59.50 & 50.74 & \textbf{91.22} & \textbf{86.99} & 72.29 \\
Voxel-fair  & 91.32 & 66.19 & 52.57 & 55.48 & 42.22 & 42.03 & 30.88 & 88.98 & 82.41 & 57.60 \\
Ray-fair  & 91.40 & 86.31 & 83.22 & 55.93 & 55.55 & 55.67 & 53.39 & 91.03 & 82.77 & 70.48 \\
BEV-fair  & 90.61 & \textbf{87.15} & \textbf{84.71} & 66.68 & \textbf{66.66} & \textbf{66.37} & \textbf{65.06} & 90.33 & 79.01 & \textbf{75.75}\\
\bottomrule 
\end{tabular}}
\label{tab:KITTI-C_fusion_model_evaluation}
\end{table*}

\subsection{Evaluation Results against Weather Corruption}
\begin{figure}[tb]
\centering
\includegraphics[width=1\linewidth]{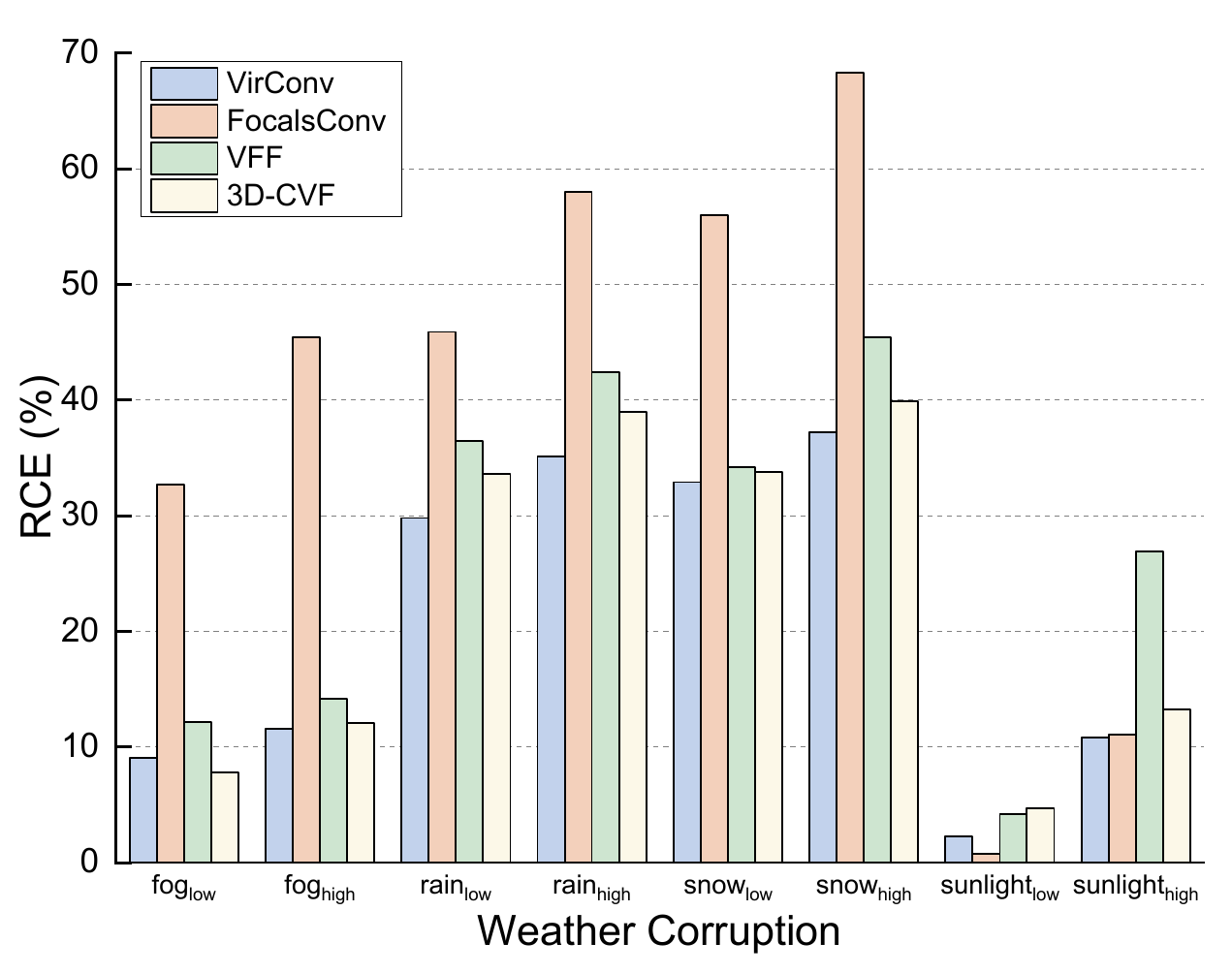}
\caption{RCE results of SOTA LiDAR-camera fusion models across four fusion modals on KITTI-C dataset.}
\label{fig:Multi-modal_SOTA_RCE}
\end{figure}

\begin{figure}[tb]
\centering
\includegraphics[width=1\linewidth]{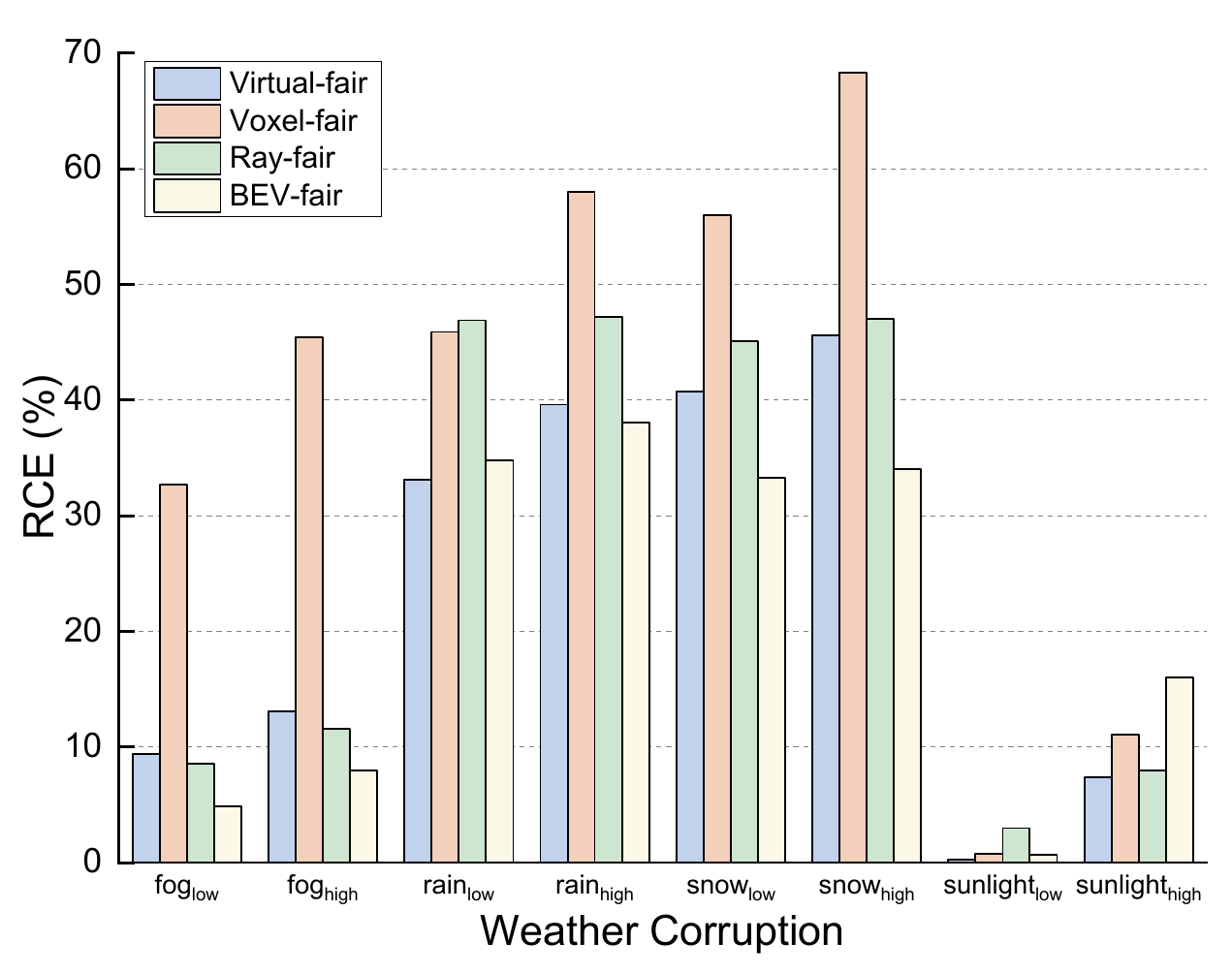}
\caption{RCE results of fair LiDAR-camera fusion models across four fusion modals on KITTI-C dataset.}
\label{fig:Multi-modal_fair_RCE}
\end{figure}

The evaluation results of LiDAR-camera fusion models against four weather corruption are shown in Table~\ref{tab:KITTI-C_fusion_model_evaluation}. In the first row, we list the weather corruption types. In the first column, we list the LiDAR-camera fusion models. In each cell, there shows the value of the AP (R40) metric. From the 3rd row to the 10th row, we demonstrate the evaluation results of eight models across different weather corruption under the 3D AP (R40) metric. Similarly, from the 12th row to the 20th row, we demonstrate the evaluation results under BEV AP (R40) metric. We also show the relative corruption error (RCE) of these models against different weather corruption in Figure~\ref{fig:Multi-modal_SOTA_RCE} and \ref{fig:Multi-modal_fair_RCE}. 

From Table~\ref{tab:KITTI-C_fusion_model_evaluation}, Figure~\ref{fig:Multi-modal_SOTA_RCE} and \ref{fig:Multi-modal_fair_RCE}, we can achieve the following conclusions. \ding{182} For all the fusion models, the performance reduction patterns are different across four weather corruption. \ding{183} Facing a specific weather corruption, the four fusion models show similar performance reduction patterns. \ding{184} Across four fusion modals, the robustness of the virtual point-based fusion model is the best.   

The detailed analysis is shown in the following paragraphs.

\noindent\textbf{Comparison from weather corruption.} \label{sec:evalustion_comparison_weathers}
\ding{182} We can easily observe that for all the eight models, the higher the corruption severity, the worse the detection performance (\eg, the performance reduction on high corruption severity is 31.3\% on average while reduction on low corruption severity is 24.2\%). 
\ding{183} Among the four weather corruption, the \textbf{rain and snow corruption} are similar and have the biggest influence on the fusion models, leading to a more than 30\% AP reduction across all eight models, showing the greatest threat across four weather corruption for 3D object detection. 
\ding{184} The \textbf{sunlight corruption} has a minor impact on the fusion models. Most of the models still maintain more than 80\% AP. This is due to that sunlight (although not strong) commonly existed in the clean dataset and has been learned by all eight models.
\ding{185} Most of the models are also affected by \textbf{fog corruption}, but the fog does not affect the models as severely as rain and snow, achieving an average AP reduction of about 14.88\%.

To sum up, the robustness of fusion models against weather conditions is intricate. The robustness of a particular model varies under different weather corruption. Additionally, models employing distinct fusion modals exhibit varied robustness under identical weather scenarios.

\noindent\textbf{Comparison from fusion modals.}\label{sec:evalustion_comparison_fusion_modals} 
\ding{182} For comparison between \textbf{SOTA} fusion models, Virtual-SOTA achieves the best performance under almost all weather corruption. Voxel-SOTA is the worst and suffers a large AP (both 3D AP and BEV AP) drop facing fog, rain, and snow corruption (RCE exceeds 50\% under snow). The robustness of Ray-SOTA is moderate and also varies greatly across different weather corruption, at an AP reduction from 0.26\% (sunlight) to 38.69\% (snow), with RCE reduction from 4.2\% to 45.4\%. The robustness of BEV-SOTA seems no good according to AP. However, due to its lower clean AP compared to the others, we think its RCE is more objective. The average RCE is only 23.0\%, which is just a bit worse than VirConv.
\ding{183} For comparison between \textbf{fair} fusion models, we can find that the Virtual-fair and BEV-fair have better robustness than others, co-occupying the highest AP values across all the weather corruption. The Voxel-fair has the worst robustness, with a significant gap (about 10\% AP on average) between it and the other three methods. Ray-fair is the moderate one.   

To sum up, considering \textbf{across all four fusion modals}, these results show that the robustness of different fusion modals when confronted with weather corruption varies greatly. The virtual point-based model has the best robustness, which may be because the virtual point generated by the image is rich both in semantic and depth information \cite{penet}. The BEV-based model is also good and just slightly worse than the virtual point-based model.


\subsection{Influence of Corrupted Single Modal}\label{sec:single-modal_evaluation}

\begin{table*}[tb]
\centering
\caption{The 3D AP results of object detectors with corrupted branches on the car category of KITTI-C at medium difficulty.}
\resizebox{1\linewidth}{!}{
\begin{tabular}{l|c|cccccccc|c}
\toprule 
model & \multicolumn{1}{c|}{clean} & fog$_{low}$ & fog$_{high}$ & rain$_{low}$ & rain$_{high}$ & snow$_{low}$ & snow$_{high}$ & sunlight$_{low}$ & sunlight$_{high}$ & average\\
\midrule 
 & \multicolumn{10}{c}{Corruption on LiDAR branch only}\\
\midrule 
Virtual-SOTA (VirConv) & \textbf{87.92} & \textbf{85.29} & \textbf{85.12} & \textbf{77.43} & \textbf{76.98} & \textbf{78.94} & \textbf{77.13} & \textbf{87.38} & 80.69 & \textbf{81.12} \tabularnewline
Voxel-SOTA (FocalsConv) & 85.52 & 74.96 & 72.44 & 65.73 & 65.69 & 65.76 & 59.21 & 85.20 & \textbf{84.57} & 71.46\\
Ray-SOTA (VFF) & 85.26 & 78.97 & 78.69 & 58.00 & 57.92 & 57.65 & 56.46 & 82.09 & 63.19 & 66.62\\
BEV-SOTA (3D-CVF) & 78.92 & 73.64 & 69.70 & 52.94 & 49.03 & 52.99 & 48.48 & 77.22 & 70.53 & 61.70\\
\midrule 
Virtual-fair  & \textbf{85.78} & \textbf{84.95} & \textbf{84.73} & \textbf{76.83} & \textbf{76.69} & \textbf{77.04} & \textbf{76.82} & \textbf{85.56} & 84.30 & \textbf{80.87} \\
Voxel-fair  & 85.52 & 74.96 & 72.44 & 65.73 & 65.69 & 65.76 & 59.21 & 85.20 & \textbf{84.57} & 71.46 \\
Ray-fair  & 85.10 & 81.71 & 79.77 & 51.82 & 51.77 & 51.74 & 50.75 & 82.83 & 73.83 & 65.53\\
BEV-fair  & 82.45 & 78.49 & 75.97 & 54.99 & 54.83 & 55.00 & 54.77 & 82.36 & 69.46 & 65.73\\
\midrule 
 & \multicolumn{10}{c}{Corruption on camera branch only}\\
\midrule 
Virtual-SOTA (VirConv) & \textbf{87.92} & \textbf{84.96} & \textbf{82.94} & \textbf{83.93} & \textbf{82.86} & \textbf{82.75} & \textbf{82.53} & \textbf{87.18} & \textbf{85.67} & \textbf{84.10}\\
Voxel-SOTA (FocalsConv) & 85.52 & 61.62 & 61.46 & 75.60 & 63.83 & 68.42 & 64.27 & 83.26 & 79.31 & 69.95 \\
Ray-SOTA (VFF) & 85.26 & 80.36 & 80.30 & 82.34 & 80.62 & 81.94 & 80.55 & 81.19 & 81.11 & 81.05 \\
BEV-SOTA (3D-CVF) & 78.92 & 77.88 & 75.71 & 78.04 & 77.91 & 78.06 & 77.89 & 77.01 & 75.30 & 77.23 \\
\midrule 
Virtual-fair  & \textbf{85.78} & 81.13 & 80.53 & 83.27 & \textbf{82.65} & 82.94 & 80.71 & \textbf{85.50} & \textbf{83.24} & 82.50 \\
Voxel-fair  & 85.52 & 61.61 & 61.46 & 75.60 & 63.83 & 68.42 & 64.27 & 83.26 & 79.31 & 69.95 \\
Ray-fair  & 85.10 & \textbf{83.21} & \textbf{83.17} & \textbf{83.32} & 80.21 & \textbf{83.30} & \textbf{83.22} & 82.16 & 82.10 & \textbf{82.60}\\
BEV-fair  & 82.45 & 82.35 & 82.34 & 82.44 & 82.42 & 82.34 & 82.24 & 82.29 & 82.11 & 82.32\\
\bottomrule 
\end{tabular}}
\label{tab:KITTI-C_single-modal_evaluation}
\end{table*}
All the LiDAR-camera fusion models have two branches (LiDAR branch and camera branch) to extract features of point cloud and image from LiDAR and camera respectively. \textbf{For the sake of brevity, here we mention that the corruption on the LiDAR or camera branch refers to the corruption on the point cloud or camera.} Considering the significance of LiDAR and camera branches as primary sources for fusion, the influence of weather corruption on \textit{single-branch} of the LiDAR-camera fusion model has not been explored yet. Thus we have a question that \textit{to what extent does each corrupted branch impact the robustness of the model?} This may help to provide ideas for improving the robustness of the fusion model against weather corruption.

We design experiments by putting the weather-corrupted data on only LiDAR/camera branch while the camera/LiDAR branch obtains the clean data and evaluates the performance of the fusion model. The results are shown in Table~\ref{tab:KITTI-C_single-modal_evaluation}. In the first row, we list the weather corruption types. In the first column, we list the fusion models. The table is divided into two parts (corruption on LiDAR-only and corruption on camera-only). In each cell, there shows the value of the 3D AP (R40) metric. Here we only show the 3D AP (R40) metric since it is the most popular and representative metric and the result of BEV AP (R40) metric is in the Section \ref{sec:more_results}. It can be seen that the effect of weather corruption on the model is generally consistent with the performance in Table~\ref{tab:KITTI-C_fusion_model_evaluation}, but corruption on LiDAR and camera branch affects the model differently.

\noindent\textbf{Comparison between corruption on LiDAR branch and camera branch.} \ding{182} Compared with applying corruption on both LiDAR and camera branches (in Table~\ref{tab:KITTI-C_fusion_model_evaluation}), the corruption on single-branch is less harmful. This suggests that when confronted with imperfect single-modal inputs, fusion methods can leverage information from another modal to enhance features and predict the final outputs. However, the influence on the robustness of the fusion model is still obvious (the AP reduction is 34.4\% at most). \ding{183} Comparing the performance of the fusion model under LiDAR/camera corruption only, we can find that the AP of the model fluctuates more significantly when subjected to corruption on the LiDAR branch while exhibiting better robustness when faced with corruption on the camera branch. For instance, the AP of the fusion model significantly decreases when exposed to weather corruption like fog, rain, and snow on the LiDAR branch (at AP reduction from 0.8\% to 34.4\%, average 15.4\%). In contrast, the decrease in AP due to weather corruption on the camera branch is less noticeable (at AP reduction from 0.1\% to 21.3\%, average 7.8\%). This means, in the fusion model, the corruption on the point cloud will bring more influence on the robustness of the fusion model than that on the image.

\noindent\textbf{Comparison between fusion modals.} \ding{182} Similar to Table~\ref{tab:KITTI-C_fusion_model_evaluation}, the overall robustness of VirConv-SOTA remains the best across SOTA-series models when subjected to weather corruption (at AP reduction from 3.0\% to 12.4\%). For fair-series, Virtual-fair is the best on the whole. It outperforms others on average when against a corrupted LiDAR branch while just a bit worse than Ray-fair against a corrupted camera branch. 
\ding{183} BEV-based and ray-based models have severe accuracy degradation when the LiDAR information is corrupted (from 5.8\% (sunlight) to 27.9\% (snow) at average on BEV-based model, from 9.7\% (sunlight) to 31.0\% (snow) at average on ray-based model), while the corruption on camera information has little effect on these two models (from 0.5\% (rain) to 1.5\% (sunlight) at average on BEV-based model, from 2.9\% (snow) to 3.6\% (sunlight) at average on ray-based model). This means the models of these two fusion modals have a high dependence on the LiDAR information and the information from the camera is not well exploited. \ding{184} Voxel-based model suffers from both corruption on LiDAR and camera branch significantly (from 0.6\% (sunlight) to 23.0\% (snow) at average with corrupted point cloud information, from 4.2\% (sunlight) to 19.2\% (snow) at average with corrupted image information). Although voxel-based model suffers from both corruption on LiDAR and camera branches, we find an interesting phenomenon that it is better than BEV-based and ray-based when only faces corruption on LiDAR branch. We can also find that its AP drop against corruption on LiDAR branch or camera branch is similar. This means voxel-based model exploits LiDAR and camera information in a balanced way, just like the virtual point-based model. 

To sum up, we find there are two points to note. (1) Most of the fusion models (except voxel-based) depend more on the information from LiDAR for detection, \ie, when faces weather corrupted point cloud, their robustness is not as good as faces weather corrupted image. (2) BEV-based and ray-based models have extremely unbalanced utilization of LiDAR and camera information, while virtual point-based and voxel-based models are relatively balanced.

\begin{figure}[tb]
\centering
    \subfigure[LiDAR points]{
    \begin{minipage}[tb]{\linewidth}
    \includegraphics[width=1\linewidth]{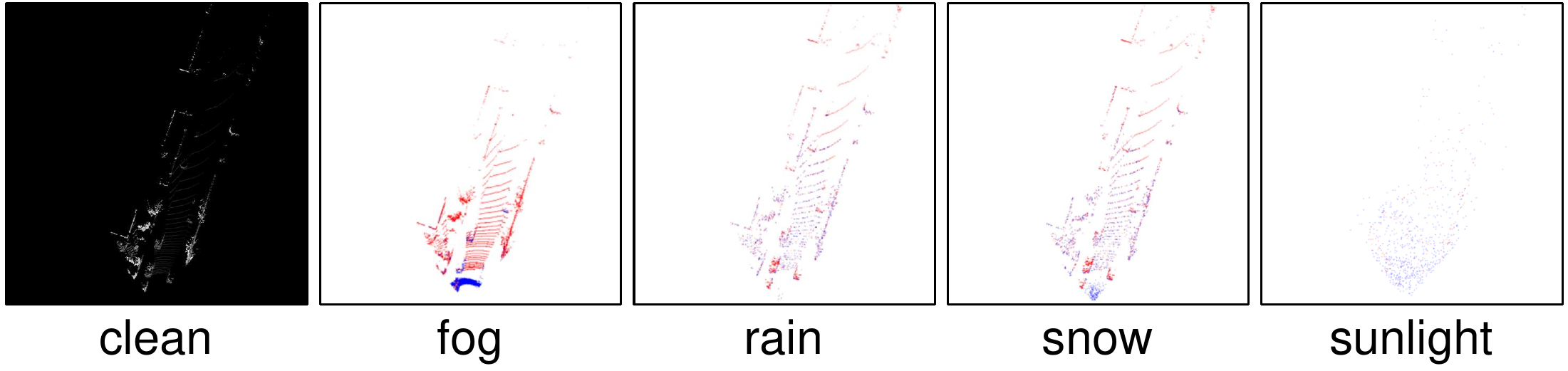}
    \end{minipage}
    }\vspace{-1mm}
    \subfigure[Virtual points generated from image]{
    \begin{minipage}[tb]{\linewidth}
    \includegraphics[width=1\linewidth]{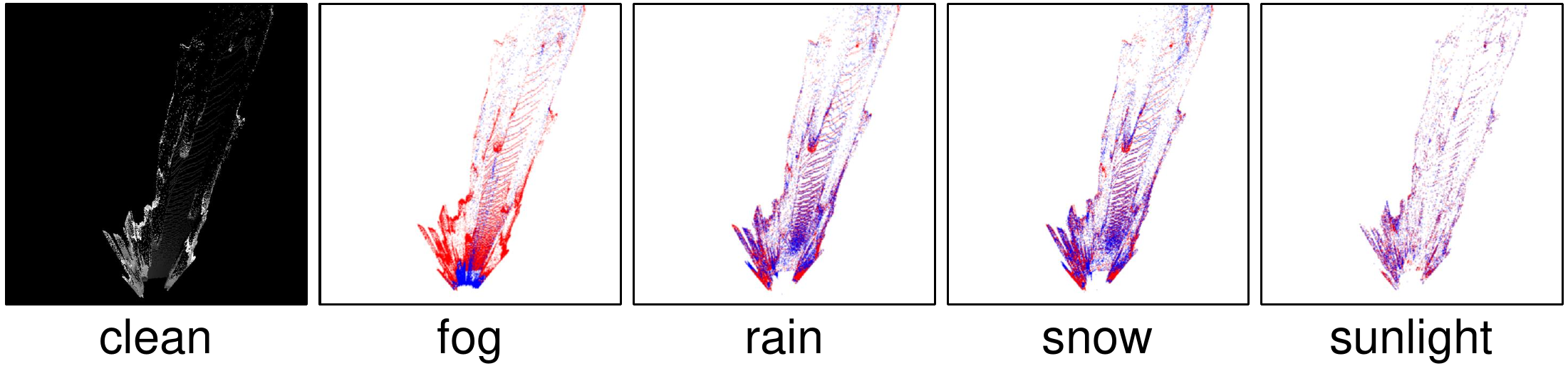}
    \end{minipage}
    }
\caption{Comparison between clean and corrupted LiDAR/camera information respectively across four high-level weather corruption. Red points are from clean information while blue points are from weather corruption.}
\label{fig:visualization_fusion_model_branches}
\end{figure}

\subsection{Visualization Analysis across Different Weather}
Table~\ref{tab:KITTI-C_fusion_model_evaluation} and~\ref{tab:KITTI-C_single-modal_evaluation} have demonstrated the quantitative robustness of fusion models under different weather corruption. To further understand the effect comprehensively, we visualize the degradation of different weather corruption on LiDAR and camera branches. Since different fusion modals have similar robustness bias under specific weather corruption, here we aim to take one of them as an example to demonstrate the visualization results. In our observation, we find only the LiDAR and camera information in the virtual point-based model can be extracted to show a satisfactory visualization result (dense and clear). 

As shown in Figure~\ref{fig:visualization_fusion_model_branches}, we show the comparison between clean and corrupted LiDAR/camera information respectively across four weather corruption under BEV angle (BEV angle is suitable for visualization). Please note that in the virtual point-based model, the camera images are transformed into virtual points. Thus we use virtual points and LiDAR points to represent the information of the camera and LiDAR respectively. In the first column of them, we show the view of clean data. From the second column to the fifth column, we calculate the difference between clean data and corrupted data. That is, red points mean points only exist in clean data while blue points mean points only exist in corrupted data. The points shared by both clean data and corrupted data are not shown in the figure. We find the following phenomenon. \ding{182} The virtual points are denser and contain more information than the LiDAR points. \ding{183} On foggy days, the LiDAR points lose much information on distant views while the virtual points retain much distant information. However, both LiDAR and virtual points have misleading blue points in the close view. \ding{184} On rainy and snowy days, both LiDAR and virtual points have information bias evenly in close and distant views. Also, in extremely close places, we can find a few misleading blue points in LiDAR points, which means the camera information is more credible. \ding{185} The LiDAR points are not affected noticeably in sunlight, while the virtual points are partially biased. 

To sum up, the influence of weather corruption on point clouds and images is very complex and with no unified rules. To improve the robustness of the model, we may need to dynamically adjust the dependence on LiDAR and camera information under different weather conditions. For example, we should refer more to the camera information on rainy and snowy days. On sunlight day, we should rely more on LiDAR information. On foggy days, it is complex and we should consider both LiDAR and camera information.

\section{Enhancement Method and Experiment}\label{sec:enhancement}
We aim to propose a \textit{general} and \textit{lightweight} robustness enhancement method against weather corruption. Here the \textit{general} means the method is useful across LiDAR-camera fusion models with different fusion modals. \textit{Lightweight} means the method is just a minor modification on the existing pipeline of LiDAR-camera fusion model.

Although the influence of weather corruption and robustness property of different modals are complex (from Section~\ref{sec:evaluation}), we have a basic realization that the exploitation of LiDAR information and camera information is not adequate and coordinated. Thus, compared with evenly and simply fusing the features extracted from the point cloud and images, we propose a simple yet effective idea by flexibly adjusting the fusion weight between features of the point cloud and images to better adapt to weather corruption and modals. 

\begin{figure}[t]
\centering
\includegraphics[width=1\linewidth]{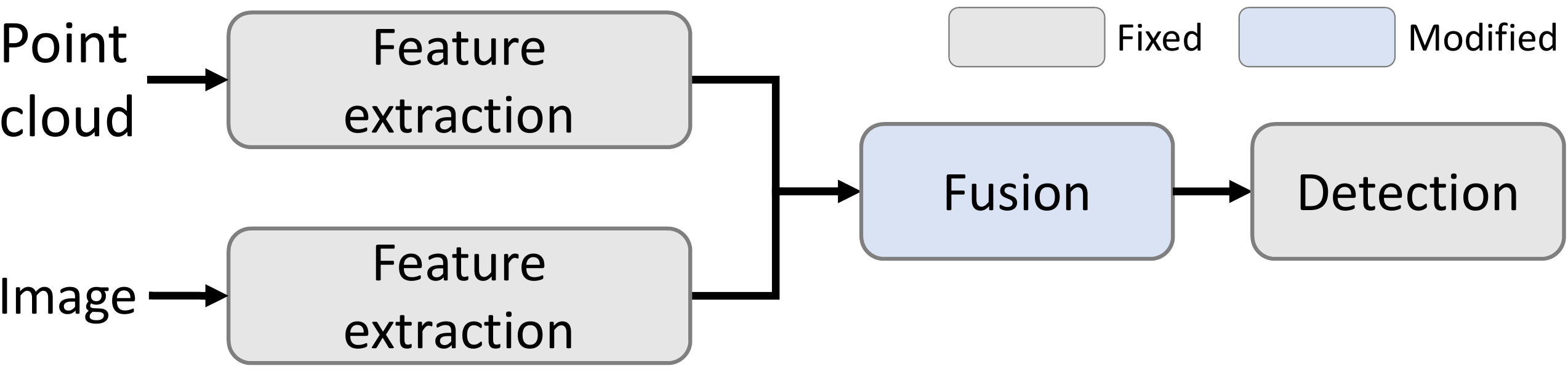}
\caption{Basic pipeline of LiDAR-camera fusion models and the fusion module is modified by our enhancement.}
\label{fig:framework}
\end{figure}

\begin{table*}[tb]
\centering
\caption{The evaluation results of object detectors with enhancement on car category of KITTI-C at medium difficulty.}
\resizebox{1\linewidth}{!}{
\begin{tabular}{l|c|cccccccc|c}
\toprule 
\multicolumn{1}{l|}{model} & \multicolumn{1}{c|}{clean} & \multirow{1}{*}{fog$_{low}$} & fog$_{high}$ & rain$_{low}$ & rain$_{high}$ & snow$_{low}$ & snow$_{high}$ & sunlight$_{low}$ & sunlight$_{high}$ & average\tabularnewline
\midrule 
Virtual-SOTA (VirConv) & 87.92 & 79.92 & 76.89 & 61.66 & 57.02 & 58.97 & 55.19 & \textcolor{blue}{85.86} & 78.42 & 69.24\tabularnewline
Virtual-Sigmoid & 87.92 & \textbf{81.08} & \textcolor{blue}{77.01} & \textcolor{blue}{61.73} & \textcolor{blue}{59.57} & \textbf{61.08} & \textcolor{blue}{55.57} & \textbf{87.47} & \textbf{85.52} & \textbf{71.13}\tabularnewline
Virtual-Attention & 87.21 & \textcolor{blue}{79.95} & \textbf{77.35} & \textbf{62.10} & \textbf{60.97} & \textcolor{blue}{60.87} & \textbf{57.51} & 85.26 & \textcolor{blue}{78.92} & \textcolor{blue}{70.37}\tabularnewline
\midrule 
Voxel-SOTA (FocalsConv) & 85.52 & 57.52 & 46.71 & 46.29 & 37.74 & 35.95 & 27.13 & \textcolor{blue}{84.79} & 76.01 & 51.52\tabularnewline
Voxel-Sigmoid & 85.62 & \textbf{59.16} & \textcolor{blue}{51.77} & \textbf{50.27} & \textbf{40.01} & \textbf{44.19} & \textbf{36.62} & \textbf{84.92} & \textcolor{blue}{76.51} & \textbf{55.43} \tabularnewline
Voxel-Attention & 85.54 & \textcolor{blue}{59.01} & \textbf{54.54} & \textcolor{blue}{47.12} & \textcolor{blue}{38.71} & \textcolor{blue}{37.68} & \textcolor{blue}{34.50} & 83.24 & \textbf{76.76} & \textcolor{blue}{53.95} \tabularnewline
\midrule
Ray-SOTA (VFF) & 85.26 & 74.85 & 73.14 & 54.15 & 49.11 & 56.09 & 46.57 & \textbf{81.66} & 62.44 & 62.25\tabularnewline
Ray-Sigmoid & 82.51 & \textcolor{blue}{76.20} & \textcolor{blue}{73.89} & \textbf{55.87} & \textcolor{blue}{51.41} & \textbf{56.96} & \textcolor{blue}{48.98} & \textcolor{blue}{80.69} & \textcolor{blue}{65.27} & \textcolor{blue}{63.66}\tabularnewline
Ray-Attention & 81.01 & \textbf{78.08} & \textbf{76.12} & \textcolor{blue}{55.30} & \textbf{55.17} & \textcolor{blue}{56.32} & \textbf{53.88} & 79.69 & \textbf{70.78} & \textbf{65.67}\tabularnewline
\midrule
BEV-SOTA (3D-CVF) & 78.92 & 72.69 & 69.43 & 52.42 & 48.14 & 52.28 & 47.43 & 75.22 & 68.41 & 60.73\tabularnewline
BEV-Sigmoid & 82.62 & \textcolor{blue}{79.06} & \textcolor{blue}{77.32} & \textcolor{blue}{54.49} & \textcolor{blue}{53.24} & \textcolor{blue}{53.78} & \textcolor{blue}{53.15} & \textcolor{blue}{81.88} & \textbf{75.60} & \textcolor{blue}{66.07}\tabularnewline
BEV-Attention & 82.52 & \textbf{80.13} & \textbf{78.10} & \textbf{55.51} & \textbf{55.60} & \textbf{55.99} & \textbf{55.15} & \textbf{82.22} & \textcolor{blue}{73.67} & \textbf{67.05}\tabularnewline
\bottomrule
\end{tabular}}
\label{tab:KITTI-C_sigmoid_attention}
\end{table*}

\begin{table}[tb]
\centering
\caption{The weight in fusion models enhanced with Sigmoid function.}
\resizebox{1\linewidth}{!}{
\begin{tabular}{l|cccccccc}
\toprule 
\multirow{2}{*}{model} & \multicolumn{2}{c}{fog} & \multicolumn{2}{c}{rain} & \multicolumn{2}{c}{snow} & \multicolumn{2}{c}{sunlight}\\
\cline{2-9}
 & $w_\mathrm{P}$ & $w_\mathrm{I}$ & $w_\mathrm{P}$ & $w_\mathrm{I}$ & $w_\mathrm{P}$ & $w_\mathrm{I}$ & $w_\mathrm{P}$ & $w_\mathrm{I}$ \\
\midrule
Virtual-Sigmoid & $\downarrow$ & $\uparrow$ & $\downarrow$ & $\uparrow$ & $\downarrow$ & $\uparrow$ & $\uparrow$ & $\downarrow$\\
Voxel-Sigmoid & $\uparrow$ & $\downarrow$ & $\uparrow$ & $\downarrow$ & $\downarrow$ & $\uparrow$ & $\uparrow$ & $\downarrow$\\
Ray-Sigmoid & $\uparrow$ & $\downarrow$ & $\downarrow$ & $\uparrow$ &  $\downarrow$& $\uparrow$ & $\uparrow$ & $\downarrow$\\
BEV-Sigmoid & $\downarrow$ & $\uparrow$ &  $\downarrow$ & $\uparrow$ & $\downarrow$ & $\uparrow$ & $\uparrow$ & $\downarrow$\\
\bottomrule
\end{tabular}
}
\label{tab:KITTI-C_sigmoid_weight}
\end{table}

\begin{figure*}[tb]
\centering
\includegraphics[width=\linewidth]{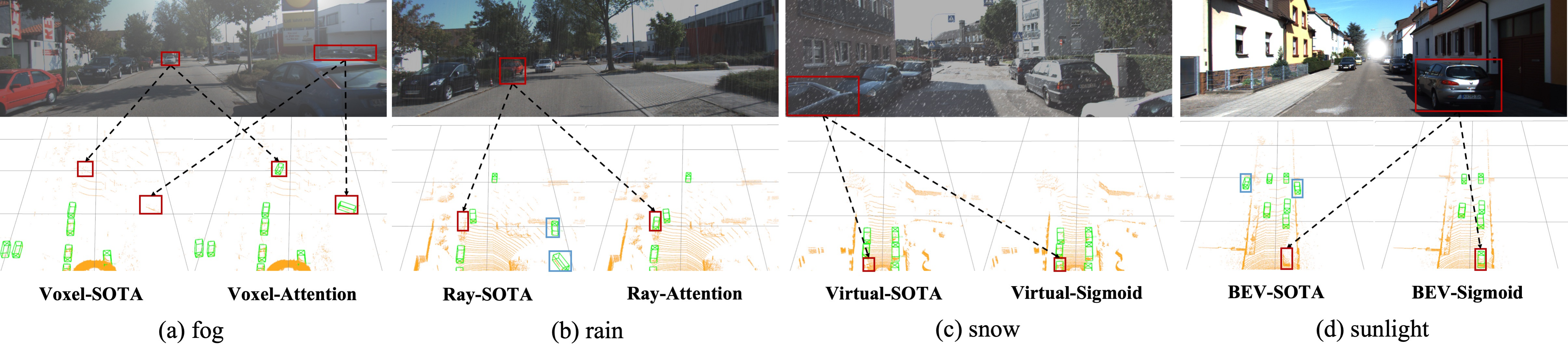}
\caption{\textbf{Visualization results.} Subfigures (a)-(d) show the comparison between the SOTA models and the enhanced ones under four weather corruption, respectively. Red boxes represent missed detections and blue boxes represent misdetections.}
\label{fig:Visualization_of_results}
\end{figure*}

\subsection{Enhancement Method}
Given a set of 3D points $\mathrm{P} = \{(x_i,y_i,z_i)\}_{i=1}^N$ and 2D images $\mathrm{I} = \{\mathrm{I}_j \in \mathbb{R}^{H \times W \times 3}\}_{j=1}^T$. The $(x_i,y_i,z_i)$ is the coordinate of the $i$-th point while $N$ is the number of points in the point cloud. $H$ and $I$ are the height and width of the input image while $T$ is the number of images. 

In Figure~\ref{fig:framework}, we only change the fusion module (in blue) of the LiDAR-camera fusion model. In the original pipeline, no matter what fusion modal, the features $\mathrm{F_{\mathrm{P}}}$ and $\mathrm{F_{\mathrm{I}}}$ are extracted from $\mathrm{P}$ and $\mathrm{I}$ respectively with the feature extraction modules. Then the features $\mathrm{F_{\mathrm{P}}}$ and $\mathrm{F_{\mathrm{I}}}$ are put into the fusion module $\mathcal{F}(\cdot,\cdot)$ and detection module $\mathcal{D}(\cdot)$ for follow-up processing. The procedure is formulated as 
\begin{align}
\mathcal{D}(\mathcal{F}(\mathrm{F_{\mathrm{P}}},\mathrm{F_{\mathrm{I}}})),
\label{eq:fusion_detection_process}
\end{align}
where \textbf{in almost all the LiDAR-camera fusion models}, the $\mathcal{F}(\cdot,\cdot)$ module just simply process the features with accumulation, \ie, $\mathcal{F}(\mathrm{F_{\mathrm{P}}},\mathrm{F_{\mathrm{I}}}) = \mathrm{F_{\mathrm{P}}}+\mathrm{F_{\mathrm{I}}}$. 
With the modified fusion module $\mathcal{\hat{F}}$, we aim to fuse features $\mathrm{F_{\mathrm{P}}}$ and $\mathrm{F_{\mathrm{I}}}$ in a more flexible way. 
To verify the universality of the proposed idea, we realize the weight adjustment in two different ways and the experiments show that both implementations are effective. To be specific, we use Sigmoid function \cite{yin2003flexible} and attention function \cite{vaswani2017attention}. \ding{182} With Sigmoid function, the fused feature $\mathrm{F}_{\mathrm{f}} = \mathcal{\hat{F}}(\mathrm{F_{\mathrm{P}}},\mathrm{F_{\mathrm{I}}}) = w_\mathrm{P} \cdot \mathrm{F_{\mathrm{P}}} + w_\mathrm{I} \cdot \mathrm{F_{\mathrm{I}}}$, where $w_\mathrm{P} = Sigmoid(MLP({\mathrm{F_{\mathrm{P}}}+\mathrm{F_{\mathrm{I}}}}))$ and $w_\mathrm{I} = 1-w_\mathrm{P}$. Here $Sigmoid(\cdot)$ means Sigmoid function and $MLP(\cdot)$ means multilayer perception. \ding{183} With attention function, the fused feature $\mathrm{F}_{\mathrm{f}} = \mathcal{\hat{F}}(\mathrm{F_{\mathrm{P}}},\mathrm{F_{\mathrm{I}}}) = Attention(\mathrm{F_{\mathrm{P}}},\mathrm{F_{\mathrm{I}}})$, where $Attention(\cdot,\cdot)$ is cross-attention function and $Query$ is $\mathrm{F_{\mathrm{I}}}$, $Key$ and $Value$ are $\mathrm{F_{\mathrm{P}}}$.

\subsection{Enhancement Results}
We show the results of enhancement methods based on Sigmoid function and Attention function in Table~\ref{tab:KITTI-C_sigmoid_attention}. We only enhance the fusion module in all SOTA models while maintaining other modules unchanged. The training procedure is the same as in Section~\ref{sec:evaluation}. In the first row, we list the weather corruption types. In the first column, we list the LiDAR-camera fusion models. In each cell, there shows the value of the 3D AP metric. The highest AP value across same-modal models is in bold while the second is in blue. The experiment results of BEV AP metrics are in Section \ref{sec:more_results}. From Table~\ref{tab:KITTI-C_sigmoid_attention}, we can find that, the AP of enhanced models on clean test datasets basically do not reduce, sometimes even become better. Sigmoid-based enhanced models achieve 2.14\%, 3.91\%, 1.41\%, and 4.69\% average AP improvement on virtual point-based, voxel-based, ray-based, and BEV-based models respectively. Also, we can find that, Attention-based enhanced models achieve 1.13\%, 2.43\%, 3.42\%, and 5.67\% AP improvement on virtual point-based, voxel-based, ray-based, and BEV-based models respectively. 

In Figure~\ref{fig:Visualization_of_results}, we visualize the comparison between SOTA models and enhanced models across four weather corruption under four scenes. The objects in red boxes and blue boxes are undetected ones and false detections by SOTA models respectively. More visualizations are shown in Section \ref{sec:visualization_entire}.

In summary, the enhancement methods proposed by us outperform the corresponding SOTA ones under different fusion modals and weather corruption. 

\noindent\textbf{Weight Analysis.}
Among two implementations, we can get the weight $w_{P}$ and $w_{I}$ of Sigmoid-based models while weight in Attention-based models can not be achieved. Thus we show the trend of weight change of Sigmoid-based models against weather corruption in Table~\ref{tab:KITTI-C_sigmoid_weight} (combine $low$ and $high$ level weather together since the trend is the same). The $\uparrow$ and $\downarrow$ mean the average weight on all corrupted test samples is larger and smaller than that facing clean test samples respectively. Please observe together with Table~\ref{tab:KITTI-C_single-modal_evaluation}. \ding{182} To achieve better robustness, we expect fusion models to depend more on the camera branch when against rain and snow (there is minor AP reduction facing corrupted image), which is consistent with Table~\ref{tab:KITTI-C_sigmoid_weight}. \ding{183} For fog weather, there is no obvious rule on whether to rely on a camera or LiDAR branch. Thus we expect the weights of fusion models to be adaptable with different modals, which is consistent with Table~\ref{tab:KITTI-C_sigmoid_weight}. \ding{184} For sunlight, the AP reduction is both small facing corrupted LiDAR or camera branch. However according to Figure~\ref{fig:visualization_fusion_model_branches}, we can find that the LiDAR branch is less influenced by sunlight and is better than the camera branch. Thus relying on LiDAR branch is more credible, which is consistent with Table~\ref{tab:KITTI-C_sigmoid_weight}. 

To sum up, with flexibly weighted fusion, the new models adjust the importance of features from point and image more intelligently against weather corruption. 


\section{Weather Corruption Details}\label{sec:weather_details}
In the context of point cloud and image simulation across different weather corruption, the methodology proposed in reference \cite{hahner2021fog} is employed across all three benchmark datasets of fog, rain, and snow.

\textbf{Fog.} 
The parameter $\alpha$, as introduced in \cite{hahner2021fog}, serves as a representative metric for meteorological optical range in authentic foggy conditions. Consistent with the conventions outlined in their work, we configure $\alpha$ to assume values of \{0.005, 0.06\} to simulate \{$low$, $high$\} levels of fog on point cloud. For the simulation of fog in the context of images, the imagaug library \cite{imgaug} is utilized, adhering to pre-defined severity levels \{$low$, $high$\} to emulate diverse intensities of fog. Additionally, a \{10\%, 50\%\}-opacity gray mask layer is introduced to further enhance the realism of the simulated atmospheric conditions.

\textbf{Rain.}
In the context of point cloud simulation for the KITTI-C dataset, we adopt the rain simulation method proposed by LISA, as delineated in reference \cite{hahner2022lidar}. We configure the parameter representing rainfall rate within LISA to assume values of \{0.2, 7.3\} to simulate \{$low$, $high$\} levels of rain. This facilitates the simulation of varying intensities of rain, spanning light rain, moderate rain, and heavy rain scenarios. For the simulation of rain effects in the context of images, we leverage the RainLayer in the imgaug library \cite{imgaug}. Herein, we set the parameter governing rainfall density to \{0.01, 0.20\}, enabling the simulation of \{$low$, $high$\} levels of rain severity. This approach ensures a nuanced representation of rain conditions across both point cloud and image modalities.

\textbf{Snow.} 
In the context of point cloud simulation, the KITTI-C dataset is subjected to snow simulation through the utilization of the LISA method \cite{kilic2021lidar}. The snowfall rate is parameterized with values of \{0.20, 7.29\}, corresponding to \{$low$, $high$\} severities for both LISA \cite{kilic2021lidar} and the approach presented in \cite{hahner2022lidar}. In the domain of image simulation, the imgaug library \cite{imgaug} is employed to emulate snow effects. The predefined severities \{$low$, $high$\} are assigned to simulate varying intensities of snowfall. To maintain consistency with the STF dataset \cite{stf}, a 30\%-opacity gray mask layer is incorporated, and the brightness is reduced by 30\%. This ensures a cohesive representation of snow conditions across both LiDAR and camera modalities.

\textbf{Strong sunlight.}
For LiDAR simulation, following observations in \cite{libre}, we simulate strong sunlight by introducing 2-meter Gaussian noise to the point cloud. We define the intensity using \{1\%, 5\%\} ratio of noisy points and categorize it into \{$low$, $high$\} levels. Similarly, for the camera simulation, we configure a {30, 70}-pixel size sun in the automold library \cite{Automold} to emulate strong sunlight of \{$low$, $high$\} levels.

\begin{figure*}[t]
\centering
\includegraphics[width=1\linewidth]{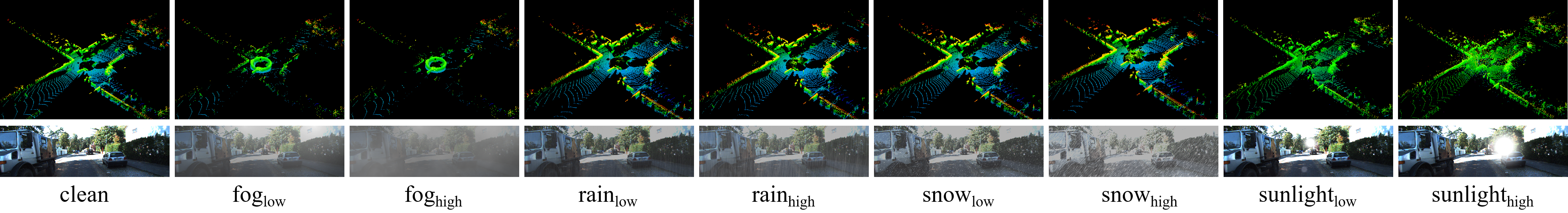}
\caption{Full visualization results of all corruption.}
\label{fig:Visualization_of_weather_corruption}
\end{figure*}

All weather types as well as his two severity levels of LiDAR and camera visualizations are listed in Figure~\ref{fig:Visualization_of_weather_corruption}.

\section{More Experiment Results}\label{sec:more_results}
In Table~\ref{tab:KITTI-C_fusion_model_evaluation_easy}, we show the AP evaluation results of the fusion models on the car category of KITTI-C under easy difficulty, as supplementary to Table~\ref{tab:KITTI-C_fusion_model_evaluation}.

In Table~\ref{tab:KITTI-C_fusion_model_evaluation_hard}, we show the AP evaluation results of the fusion models on the car category of KITTI-C under hard difficulty, as supplementary to Table~\ref{tab:KITTI-C_fusion_model_evaluation}.

In Table~\ref{tab:KITTI-C_single-modal_evaluation_BEV_AP}, we show the BEV AP evaluation results of the fusion models with a single corrupted branch on the car category of KITTI-C under medium difficulty, as supplementary to Table~\ref{tab:KITTI-C_single-modal_evaluation}.

In Table~\ref{tab:KITTI-C_single-modal_evaluation_3D_AP_easy}, we show the 3D AP evaluation results of the fusion models with a single corrupted branch on the car category of KITTI-C under easy difficulty, as supplementary to Table~\ref{tab:KITTI-C_single-modal_evaluation}.

In Table~\ref{tab:KITTI-C_single-modal_evaluation_BEV_AP_easy}, we show the BEV AP evaluation results of the fusion models with a single corrupted branch on the car category of KITTI-C under easy difficulty, as supplementary to Table~\ref{tab:KITTI-C_single-modal_evaluation}.

In Table~\ref{tab:KITTI-C_single-modal_evaluation_3D_AP_hard}, we show the 3D AP evaluation results of the fusion models with a single corrupted branch on the car category of KITTI-C under hard difficulty, as supplementary to Table~\ref{tab:KITTI-C_single-modal_evaluation}.

In Table~\ref{tab:KITTI-C_single-modal_evaluation_BEV_AP_hard}, we show the BEV AP evaluation results of the fusion models with a single corrupted branch on the car category of KITTI-C under hard difficulty, as supplementary to Table~\ref{tab:KITTI-C_single-modal_evaluation}.

In Table~\ref{tab:KITTI-C_sigmoid_attention_mid_BEV}, we show the BEV AP evaluation results of the fusion models and their enhanced methods under medium difficulty, as supplementary to Table~\ref{tab:KITTI-C_sigmoid_attention}.

In Table~\ref{tab:KITTI-C_sigmoid_attention_3D_easy}, we show the 3D AP evaluation results of the fusion models and their enhanced methods under easy difficulty, as supplementary to Table~\ref{tab:KITTI-C_sigmoid_attention}.

In Table~\ref{tab:KITTI-C_sigmoid_attention_easy_BEV}, we show the BEV AP evaluation results of the fusion models and their enhanced methods under easy difficulty, as supplementary to Table~\ref{tab:KITTI-C_sigmoid_attention}.

In Table~\ref{tab:KITTI-C_sigmoid_attention_3D_hard}, we show the 3D AP evaluation results of the fusion models and their enhanced methods under hard difficulty, as supplementary to Table~\ref{tab:KITTI-C_sigmoid_attention}.

In Table~\ref{tab:KITTI-C_sigmoid_attention_hard_BEV}, we show the BEV AP evaluation results of the fusion models and their enhanced methods under hard difficulty, as supplementary to Table~\ref{tab:KITTI-C_sigmoid_attention}.
\begin{table*}[tb]
\centering
\caption{The evaluation results of eight different 3D object detectors on car category of KITTI-C at easy difficulty.}
\resizebox{1\linewidth}{!}{
\begin{tabular}{l|c|cccccccc|c}
\toprule 
model & clean & fog$_{low}$ & fog$_{high}$ & rain$_{low}$ & rain$_{high}$ & snow$_{low}$ & snow$_{high}$ & sunlight$_{low}$ & sunlight$_{high}$ & average \\ \midrule 
 & \multicolumn{10}{c}{Car 3D AP (R40)}\\ \midrule
Virtual-SOTA (VirConv) & \textbf{92.77} & \textbf{91.10} & \textbf{90.79} & \textbf{82.00} & \textbf{79.50} & \textbf{79.58} & \textbf{77.43} & \textbf{92.20} & 81.37 & \textbf{84.25}\\ 
Voxel-SOTA (FocalsConv) & 92.47 & 80.24 & 72.63 & 63.72 & 57.20 & 55.08 & 41.51 & 92.17 & \textbf{86.99} & 69.07\\ 
Ray-SOTA (VFF) & 89.62 & 86.71 & 85.73 & 74.08 & 69.72 & 75.87 & 66.64 & 88.32 & 63.81 & 76.36\\ 
BEV-SOTA (3D-CVF) & 89.17 & 84.01 & 82.79 & 74.98 & 71.58 & 76.05 & 68.88 & 84.28 & 68.91 & 76.44 \\ \midrule
Virtual-fair  & \textbf{92.77} & \textbf{91.55} & \textbf{88.31} & \textbf{77.61} & 72.19 & \textbf{74.18} & 70.34 & \textbf{92.62} & \textbf{89.08} & \textbf{81.99}\\
Voxel-fair  & 92.47 & 80.24 & 72.63 & 63.72 & 57.20 & 55.08 & 41.51 & 92.17 & 86.99 & 69.07 \\
Ray-fair  & 92.12 & 88.90 & 88.19 & 65.34 & 65.05 & 69.43 & 65.43 & 91.45 & 77.42 & 76.40 \\
BEV-fair  & 92.26 & 88.65 & 88.17 & 74.87 & \textbf{74.73} & 73.21 & \textbf{73.18} & 90.79 & 71.94 & 79.82\\
\midrule 
 & \multicolumn{10}{c}{Car BEV AP (R40)}\\
\midrule 
Virtual-SOTA (VirConv) & \textbf{95.96} & \textbf{94.29} & \textbf{92.93} & \textbf{87.01} & \textbf{85.06} & \textbf{86.21} & \textbf{84.06} & \textbf{95.37} & 85.67 & \textbf{88.82}\\
Voxel-SOTA (FocalsConv) & 93.22 & 87.90 & 80.03 & 73.40 & 64.81 & 62.38 & 46.77 & 93.15 & \textbf{91.07} & 74.94\\
Ray-SOTA (VFF) & 92.77 & 91.38 & 90.34 & 83.02 & 79.89 & 84.38 & 77.66 & 91.61 & 70.74 & 83.63 \\
BEV-SOTA (3D-CVF) & 90.41 & 90.12 & 88.18 & 82.70 & 79.22 & 83.68 & 78.20 & 85.44 & 74.67 & 82.78 \\
\midrule 
Virtual-fair  & \textbf{95.94} & \textbf{93.44} & \textbf{92.65} & \textbf{84.54} & 79.49 & 81.48 & 77.26 & \textbf{95.90} & \textbf{92.75} & \textbf{87.19} \\
Voxel-fair  & 93.22 & 87.90 & 80.03 & 73.40 & 64.81 & 62.38 & 46.77 & 93.15 & 91.07 & 74.94 \\
Ray-fair  & 95.55 & 92.68 & 92.31 & 74.77 & 74.68 & 77.02 & 74.80 & 94.74 & 84.25 & 83.16 \\
BEV-fair  & 95.55 & 92.07 & 91.96 & 82.59 & \textbf{82.60} & \textbf{82.84} & \textbf{82.31} & 94.49 & 77.54 & 85.80\\
\bottomrule 
\end{tabular}}
\label{tab:KITTI-C_fusion_model_evaluation_easy}
\end{table*}

\begin{table*}[tb]
\centering
\caption{The evaluation results of eight different 3D object detectors on car category of KITTI-C at hard difficulty.}
\resizebox{1\linewidth}{!}{
\begin{tabular}{l|c|cccccccc|c}
\toprule 
model & clean & fog$_{low}$ & fog$_{high}$ & rain$_{low}$ & rain$_{high}$ & snow$_{low}$ & snow$_{high}$ & sunlight$_{low}$ & sunlight$_{high}$ & average \\ \midrule 
 & \multicolumn{10}{c}{Car 3D AP (R40)}\\ \midrule
Virtual-SOTA (VirConv) & \textbf{85.60} & \textbf{76.77} & \textbf{72.99} & \textbf{55.55} & \textbf{52.81} & \textbf{54.84} & \textbf{50.84} & \textbf{84.96} & \textbf{74.10} & \textbf{65.36} \\ 
Voxel-SOTA (FocalsConv) & 85.07 & 53.25 & 44.09 & 42.68 & 33.94 & 33.89 & 25.11 & 82.97 & 74.09 & 48.75\\ 
Ray-SOTA (VFF) & 80.18 & 69.60 & 66.78 & 50.72 & 44.37 & 51.46 & 42.08 & 79.57 & 60.59 & 58.15\\ 
BEV-SOTA (3D-CVF) & 78.16 & 67.44 & 63.53 & 46.14 & 43.07 & 45.79 & 41.23 & 71.70 & 65.36 & 55.53 \\ \midrule
Virtual-fair  & 84.97 & 74.54 & 70.14 & \textbf{52.96} & 44.32 & 48.08 & 42.28 & \textbf{82.95} & \textbf{76.54} & 61.48\\
Voxel-fair  & \textbf{85.07} & 53.25 & 44.09 & 42.68 & 33.94 & 33.89 & 25.11 & 82.97 & 74.09 & 48.75 \\
Ray-fair  & 82.87 & 74.44 & 70.61 & 41.06 & 41.03 & 41.04 & 40.67 & 82.18 & 70.05 & 57.64 \\
BEV-fair  & 80.39 & \textbf{75.56} & \textbf{73.13} & 51.74 & \textbf{51.88} & \textbf{51.83} & \textbf{49.96} & 79.84 & 67.93 & \textbf{62.73} \\
\midrule 
 & \multicolumn{10}{c}{Car BEV AP (R40)}\\
\midrule 
Virtual-SOTA (VirConv) & \textbf{91.33} & \textbf{83.53} & \textbf{81.58} & \textbf{64.27} & \textbf{60.72} & \textbf{63.47} & \textbf{57.16} & \textbf{89.26} & \textbf{80.78} & \textbf{72.60}\\
Voxel-SOTA (FocalsConv) & 91.10 & 61.89 & 48.06 & 51.39 & 37.92 & 37.76 & 28.67 & 88.34 & 80.53 & 54.32\\
Ray-SOTA (VFF) & 88.23 & 80.22 & 77.62 & 61.88 & 55.70 & 62.71 & 53.14 & 87.69 & 71.67 & 68.83 \\
BEV-SOTA (3D-CVF) & 87.60 & 77.05 & 72.83 & 54.75 & 49.88 & 54.09 & 47.39 & 78.32 & 69.31 & 62.95 \\
\midrule 
Virtual-fair  & 89.10 & 81.74 & 78.76 & \textbf{63.86} & 50.55 & 54.79 & 46.09 & \textbf{88.83} & \textbf{84.52} & 68.64 \\
Voxel-fair  & \textbf{91.10} & 61.89 & 48.06 & 51.39 & 37.92 & 37.76 & 28.67 & 88.34 & 80.53 & 54.32 \\
Ray-fair  & 89.16 & 83.37 & 79.25 & 51.54 & 45.15 & 49.77 & 48.92 & 88.75 & 80.59 & 65.92 \\
BEV-fair  & 88.79 & \textbf{84.78} & \textbf{82.28} & 62.94 & \textbf{62.92} & \textbf{62.69} & \textbf{61.16} & 88.54 & 78.88 & \textbf{73.02} \\
\bottomrule 
\end{tabular}}
\label{tab:KITTI-C_fusion_model_evaluation_hard}
\end{table*}

\begin{table*}[t]
\centering
\caption{The BEV AP evaluation results of object detectors with corrupted branches on car category of KITTI-C at medium difficulty.}
\resizebox{1\linewidth}{!}{
\begin{tabular}{l|c|cccccccc|c}
\toprule 
model & \multicolumn{1}{c|}{clean} & fog$_{low}$ & fog$_{high}$ & rain$_{low}$ & rain$_{high}$ & snow$_{low}$ & snow$_{high}$ & sunlight$_{low}$ & sunlight$_{high}$ & average\\
\midrule 
 & \multicolumn{10}{c}{Corruption on LiDAR branch only}\\
\midrule 
Virtual-SOTA (VirConv) & \textbf{91.71} & \textbf{91.22} & \textbf{91.18} & \textbf{86.38} & \textbf{86.36} & \textbf{86.54} & \textbf{86.37} & \textbf{91.56} & 85.59 & \textbf{88.15} \tabularnewline
Voxel-SOTA (FocalsConv) & 91.32 & 84.83 & 82.12 & 75.76 & 75.59 & 77.39 & 62.74 & 91.31 & \textbf{91.01} & 80.09\\
Ray-SOTA (VFF) & 89.23 & 86.89 & 85.98 & 71.01 & 70.70 & 71.01 & 69.40 & 86.83 & 73.86 & 76.96\\
BEV-SOTA (3D-CVF) & 88.20 & 83.31 & 79.12 & 61.89 & 57.45 & 62.42 & 54.93 & 81.67 & 77.03 & 71.73\\
\midrule 
Virtual-fair  & \textbf{91.52} & \textbf{91.05} & \textbf{90.72} & \textbf{85.69} & \textbf{85.74} & \textbf{85.71} & \textbf{85.69} & 91.15 & 90.15 & \textbf{88.24} \\
Voxel-fair  & 91.32 & 84.83 & 82.12 & 75.76 & 75.59 & 77.39 & 62.74 & \textbf{91.31} & \textbf{91.01} & 80.09 \\
Ray-fair  & 91.40 & 88.40 & 87.81 & 62.58 & 62.54 & 63.39 & 62.51 & 90.94 & 81.11 & 74.91\\
BEV-fair  & 90.61 & 86.96 & 84.64 & 66.49 & 66.37 & 66.35 & 65.35 & 88.39 & 79.06 & 75.45\\
\midrule 
 & \multicolumn{10}{c}{Corruption on camera branch only}\\
\midrule 
Virtual-SOTA (VirConv) & \textbf{91.71} & \textbf{91.41} & \textbf{91.01} & \textbf{90.93} & \textbf{89.05} & \textbf{90.79} & \textbf{88.83} & \textbf{91.70} & \textbf{91.55} & \textbf{90.66}\\
Voxel-SOTA (FocalsConv) & 91.32 & 67.38 & 67.35 & 83.83 & 69.65 & 76.60 & 75.48 & 88.95 & 85.13 & 76.80 \\
Ray-SOTA (VFF) & 89.23 & 88.57 & 88.44 & 88.51 & 88.75 & 88.54 & 88.30 & 88.16 & 88.13 & 88.43 \\
BEV-SOTA (3D-CVF) & 88.20 & 88.35 & 85.27 & 88.10 & 87.92 & 88.13 & 86.72 & 86.44 & 86.38 & 87.16 \\
\midrule 
Virtual-fair  & \textbf{91.52} & 89.29 & 87.00 & 89.48 & 89.21 & 89.32 & 87.13 & \textbf{91.69} & 89.18 & 89.04 \\
Voxel-fair  & 91.32 & 67.38 & 67.35 & 83.83 & 69.65 & 76.60 & 75.48 & 88.95 & 85.13 & 76.80 \\
Ray-fair  & 91.40 & 89.42 & \textbf{89.42} & 89.43 & 89.42 & 89.44 & \textbf{89.40} & 91.38 & \textbf{91.22} & \textbf{89.89}\\
BEV-fair  & 90.61 & \textbf{90.03} & 88.65 & \textbf{90.41} & \textbf{90.32} & \textbf{90.38} & 88.65 & 88.35 & 87.91 & 89.34\\
\bottomrule 
\end{tabular}}
\label{tab:KITTI-C_single-modal_evaluation_BEV_AP}
\end{table*}

\begin{table*}[t]
\centering
\caption{The 3D AP evaluation results of object detectors with corrupted branches on car category of KITTI-C at easy difficulty.}
\resizebox{1\linewidth}{!}{
\begin{tabular}{l|c|cccccccc|c}
\toprule 
model & \multicolumn{1}{c|}{clean} & fog$_{low}$ & fog$_{high}$ & rain$_{low}$ & rain$_{high}$ & snow$_{low}$ & snow$_{high}$ & sunlight$_{low}$ & sunlight$_{high}$ & average\\
\midrule 
 & \multicolumn{10}{c}{Corruption on LiDAR branch only}\\
\midrule 
Virtual-SOTA (VirConv) & \textbf{92.77} & \textbf{91.66} & \textbf{91.55} & \textbf{88.90} & \textbf{88.50} & \textbf{88.57} & \textbf{89.07} & 92.35 & 81.72 & \textbf{89.04} \tabularnewline
Voxel-SOTA (FocalsConv) & 92.47 & 89.46 & 88.59 & 81.80 & 81.78 & 81.83 & 82.23 & \textbf{92.84} & \textbf{91.14} & 86.21\\
Ray-SOTA (VFF) & 89.62 & 88.21 & 88.19 & 77.95 & 76.31 & 78.57 & 76.60 & 88.53 & 64.33 & 79.34\\
BEV-SOTA (3D-CVF) & 89.17 & 84.41 & 83.01 & 76.53 & 73.02 & 77.93 & 70.32 & 88.27 & 80.19 & 79.21\\
\midrule 
Virtual-fair  & \textbf{92.77} & \textbf{92.54} & \textbf{92.49} & \textbf{88.86} & \textbf{88.65} & \textbf{90.73} & \textbf{88.51} & 92.45 & 90.49 & \textbf{90.59} \\
Voxel-fair  & 92.47 & 89.46 & 88.59 & 81.80 & 81.78 & 81.83 & 82.23 & \textbf{92.84} & \textbf{91.14} & 86.21 \\
Ray-fair  & 92.12 & 91.13 & 88.97 & 71.88 & 71.66 & 73.06 & 72.16 & 91.51 & 77.66 & 80.01\\
BEV-fair  & 92.26 & 89.51 & 87.85 & 74.83 & 73.26 & 75.55 & 74.15 & 90.94 & 72.10 & 79.56\\
\midrule 
 & \multicolumn{10}{c}{Corruption on camera branch only}\\
\midrule 
Virtual-SOTA (VirConv) & \textbf{92.77} & \textbf{92.55} & \textbf{92.34} & \textbf{92.47} & \textbf{92.46} & \textbf{92.37} & \textbf{92.60} & \textbf{92.80} & \textbf{92.77} & \textbf{92.55}\\
Voxel-SOTA (FocalsConv) & 92.47 & 88.74 & 85.16 & 86.18 & 85.67 & 82.69 & 75.48 & 92.18 & 88.94 & 85.63 \\
Ray-SOTA (VFF) & 89.62 & 89.33 & 89.74 & 89.55 & 89.44 & 89.79 & 88.93 & 89.52 & 89.51 & 89.48 \\
BEV-SOTA (3D-CVF) & 89.17 & 89.13 & 88.76 & 89.20 & 87.81 & 88.97 & 86.33 & 89.16 & 85.33 & 88.09 \\
\midrule 
Virtual-fair  & \textbf{92.77} & \textbf{92.94} & \textbf{92.41} & \textbf{92.72} & \textbf{92.69} & \textbf{92.74} & \textbf{92.42} & \textbf{92.75} & \textbf{92.69} & \textbf{92.68} \\
Voxel-fair  & 92.47 & 88.74 & 85.16 & 86.18 & 85.67 & 82.69 & 75.48 & 92.18 & 88.94 & 85.63 \\
Ray-fair  & 92.12 & 91.95 & 92.02 & 91.88 & 91.73 & 91.88 & 91.85 & 92.15 & 92.00 & 91.93\\
BEV-fair  & 92.26 & 91.96 & 91.94 & 92.26 & 92.25 & 92.55 & 91.94 & 91.94 & 91.74 & 92.07\\
\bottomrule 
\end{tabular}}
\label{tab:KITTI-C_single-modal_evaluation_3D_AP_easy}
\end{table*}

\begin{table*}[t]
\centering
\caption{The BEV AP evaluation results of object detectors with corrupted branches on car category of KITTI-C at easy difficulty.}
\resizebox{1\linewidth}{!}{
\begin{tabular}{l|c|cccccccc|c}
\toprule 
model & \multicolumn{1}{c|}{clean} & fog$_{low}$ & fog$_{high}$ & rain$_{low}$ & rain$_{high}$ & snow$_{low}$ & snow$_{high}$ & sunlight$_{low}$ & sunlight$_{high}$ & average\\
\midrule 
 & \multicolumn{10}{c}{Corruption on LiDAR branch only}\\
\midrule 
Virtual-SOTA (VirConv) & \textbf{95.96} & \textbf{95.63} & \textbf{94.72} & \textbf{92.91} & \textbf{92.86} & \textbf{93.03} & \textbf{92.83} & \textbf{95.56} & 85.69 & \textbf{92.90} \tabularnewline
Voxel-SOTA (FocalsConv) & 93.22 & 92.73 & 91.83 & 86.97 & 89.34 & 88.74 & 87.21 & 93.43 & \textbf{92.89} & 90.39\\
Ray-SOTA (VFF) & 92.77 & 91.53 & 91.37 & 85.51 & 85.50 & 85.57 & 85.39 & 89.61 & 71.49 & 85.75\\
BEV-SOTA (3D-CVF) & 90.41 & 90.32 & 89.28 & 85.43 & 81.82 & 83.69 & 79.01 & 85.67 & 79.84 & 84.38\\
\midrule 
Virtual-fair  & \textbf{95.94} & \textbf{96.03} & \textbf{95.92} & \textbf{92.93} & \textbf{92.82} & \textbf{92.82} & \textbf{92.73} & \textbf{95.71} & \textbf{93.95} & \textbf{94.11} \\
Voxel-fair  & 93.22 & 92.73 & 91.83 & 86.97 & 89.34 & 88.74 & 87.21 & 93.43 & 92.89 & 90.39 \\
Ray-fair  & 95.55 & 92.76 & 92.63 & 79.22 & 79.03 & 73.87 & 79.06 & 92.50 & 84.50 & 84.20 \\
BEV-fair  & 95.55 & 92.14 & 92.02 & 82.69 & 82.61 & 83.08 & 82.35 & 92.49 & 77.72 & 86.89 \\
\midrule 
 & \multicolumn{10}{c}{Corruption on camera branch only}\\
\midrule 
Virtual-SOTA (VirConv) & \textbf{95.96} & \textbf{95.87} & \textbf{95.59} & \textbf{95.82} & \textbf{95.56} & \textbf{95.80} & \textbf{95.70} & \textbf{95.95} & \textbf{95.88} & \textbf{95.77}\\
Voxel-SOTA (FocalsConv) & 93.22 & 89.05 & 87.08 & 92.34 & 89.33 & 89.88 & 86.65 & 93.07 & 92.02 & 89.93 \\
Ray-SOTA (VFF) & 92.77 & 92.85 & 92.51 & 92.77 & 92.55 & 92.38 & 92.33 & 92.65 & 92.63 & 92.58 \\
BEV-SOTA (3D-CVF) & 90.41 & 90.33 & 90.31 & 89.99 & 90.01 & 90.25 & 90.12 & 90.40 & 89.77 & 90.15 \\
\midrule 
Virtual-fair  & \textbf{95.94} & \textbf{96.22} & \textbf{95.86} & \textbf{95.96} & \textbf{95.71} & \textbf{95.87} & \textbf{93.64} & 95.33 & 94.95 & \textbf{95.45} \\
Voxel-fair  & 93.22 & 89.05 & 87.08 & 92.34 & 89.33 & 89.88 & 86.65 & 93.07 & 92.02 & 89.93 \\
Ray-fair  & 95.55 & 93.36 & 93.36 & 93.37 & 93.24 & 93.25 & 93.23 & \textbf{95.51} & \textbf{95.41} & 93.84\\
BEV-fair  & 95.55 & 95.09 & 95.09 & 95.55 & 95.55 & 95.66 & 95.09 & 95.09 & 95.09 & 95.28 \\
\bottomrule 
\end{tabular}}
\label{tab:KITTI-C_single-modal_evaluation_BEV_AP_easy}
\end{table*}

\begin{table*}[t]
\centering
\caption{The 3D AP evaluation results of object detectors with corrupted branches on car category of KITTI-C at hard difficulty.}
\resizebox{1\linewidth}{!}{
\begin{tabular}{l|c|cccccccc|c}
\toprule 
model & \multicolumn{1}{c|}{clean} & fog$_{low}$ & fog$_{high}$ & rain$_{low}$ & rain$_{high}$ & snow$_{low}$ & snow$_{high}$ & sunlight$_{low}$ & sunlight$_{high}$ & average\\
\midrule 
 & \multicolumn{10}{c}{Corruption on LiDAR branch only}\\
\midrule 
Virtual-SOTA (VirConv) & \textbf{85.60} & \textbf{82.39} & \textbf{82.32} & \textbf{72.36} & \textbf{72.25} & \textbf{72.43} & \textbf{72.28} & \textbf{84.95} & 76.29 & \textbf{76.91} \tabularnewline
Voxel-SOTA (FocalsConv) & 85.07 & 72.04 & 69.29 & 63.53 & 61.97 & 63.56 & 62.56 & 83.43 & \textbf{82.50} & 69.86\\
Ray-SOTA (VFF) & 80.18 & 75.60 & 74.09 & 55.36 & 54.20 & 54.19 & 54.16 & 78.65 & 62.87 & 63.64\\
BEV-SOTA (3D-CVF) & 78.16 & 70.44 & 66.53 & 49.72 & 46.07 & 49.89 & 49.85 & 77.43 & 67.35 & 59.66\\
\midrule 
Virtual-fair  & 84.97 & \textbf{81.91} & \textbf{80.00} & \textbf{72.14} & \textbf{72.02} & \textbf{72.08} & \textbf{72.32} & \textbf{84.47} & 81.32 & \textbf{77.03} \\
Voxel-fair  & \textbf{85.07} & 72.04 & 69.29 & 63.53 & 61.97 & 63.56 & 62.56 & 83.43 & \textbf{82.50} & 69.86 \\
Ray-fair  & 82.87 & 77.28 & 75.19 & 48.12 & 48.06 & 48.98 & 48.03 & 81.94 & 70.14 & 62.22 \\
BEV-fair  & 80.39 & 76.28 & 73.89 & 52.73 & 52.63 & 52.68 & 52.42 & 80.69 & 68.00 & 63.67\\
\midrule 
 & \multicolumn{10}{c}{Corruption on camera branch only}\\
\midrule 
Virtual-SOTA (VirConv) & \textbf{85.60} & \textbf{82.65} & \textbf{80.64} & \textbf{82.55} & \textbf{80.80} & \textbf{82.37} & \textbf{80.52} & \textbf{85.51} & \textbf{85.19} & \textbf{82.53}\\
Voxel-SOTA (FocalsConv) & 85.07 & 59.49 & 57.19 & 75.61 & 61.79 & 66.52 & 57.09 & 82.95 & 77.60 & 67.28 \\
Ray-SOTA (VFF) & 80.18 & 78.02 & 77.60 & 80.16 & 78.35 & 80.01 & 78.16 & 80.10 & 80.02 & 79.05 \\
BEV-SOTA (3D-CVF) & 78.16 & 76.12 & 75.85 & 73.61 & 73.17 & 71.94 & 71.76 & 78.10 & 77.96 & 74.81 \\
\midrule 
Virtual-fair  & 84.97 & 80.39 & 78.28 & \textbf{80.98} & 80.48 & 80.72 & 78.40 & \textbf{84.36} & \textbf{82.70} & 80.79 \\
Voxel-fair  & \textbf{85.07} & 59.49 & 57.19 & 75.61 & 61.79 & 66.52 & 57.09 & 82.95 & 77.60 & 67.28 \\
Ray-fair  & 82.87 & \textbf{80.78} & \textbf{80.74} & 80.90 & \textbf{80.75} & \textbf{80.85} & \textbf{80.75} & 81.92 & 81.50 & \textbf{81.02}\\
BEV-fair  & 80.39 & 80.23 & 80.25 & 80.39 & 80.38 & 80.76 & 80.25 & 80.25 & 80.25 & 80.34\\
\bottomrule 
\end{tabular}}
\label{tab:KITTI-C_single-modal_evaluation_3D_AP_hard}
\end{table*}

\begin{table*}[t]
\centering
\caption{The BEV AP evaluation results of object detectors with corrupted branches on car category of KITTI-C at hard difficulty.}
\resizebox{1\linewidth}{!}{
\begin{tabular}{l|c|cccccccc|c}
\toprule 
model & \multicolumn{1}{c|}{clean} & fog$_{low}$ & fog$_{high}$ & rain$_{low}$ & rain$_{high}$ & snow$_{low}$ & snow$_{high}$ & sunlight$_{low}$ & sunlight$_{high}$ & average\\
\midrule 
 & \multicolumn{10}{c}{Corruption on LiDAR branch only}\\
\midrule 
Virtual-SOTA (VirConv) & \textbf{91.33} & \textbf{88.41} & \textbf{88.28} & \textbf{81.24} & \textbf{81.23} & \textbf{81.42} & \textbf{81.24} & 89.11 & 82.83 & \textbf{84.22} \tabularnewline
Voxel-SOTA (FocalsConv) & 91.10 & 80.74 & 77.91 & 73.54 & 73.40 & 73.49 & 73.28 & \textbf{91.02} & \textbf{88.89} & 79.03\\
Ray-SOTA (VFF) & 88.23 & 83.83 & 81.81 & 67.28 & 67.06 & 67.17 & 66.64 & 88.37 & 73.17 & 74.42\\
BEV-SOTA (3D-CVF) & 87.60 & 80.35 & 75.91 & 68.39 & 68.10 & 68.29 & 68.25 & 86.98 & 75.59 & 73.98\\
\midrule 
Virtual-fair  & 89.10 & \textbf{88.26} & \textbf{87.92} & \textbf{81.05} & \textbf{80.98} & \textbf{81.06} & \textbf{80.95} & 88.78 & 87.53 & \textbf{84.57} \\
Voxel-fair  & \textbf{91.10} & 80.74 & 77.91 & 73.54 & 73.40 & 73.49 & 73.28 & \textbf{91.02} & \textbf{88.89} & 79.03 \\
Ray-fair  & 89.16 & 85.65 & 83.45 & 58.74 & 58.72 & 59.49 & 58.65 & 88.69 & 80.29 & 71.71\\
BEV-fair  & 88.79 & 84.70 & 82.29 & 63.74 & 63.67 & 63.30 & 62.65 & 88.35 & 78.95 & 73.46\\
\midrule 
 & \multicolumn{10}{c}{Corruption on camera branch only}\\
\midrule 
Virtual-SOTA (VirConv) & \textbf{91.33} & \textbf{89.13} & \textbf{88.92} & \textbf{88.94} & \textbf{88.92} & \textbf{88.85} & \textbf{86.74} & \textbf{91.40} & \textbf{91.23} & \textbf{89.27}\\
Voxel-SOTA (FocalsConv) & 91.10 & 65.25 & 62.96 & 82.01 & 67.62 & 74.70 & 60.56 & 88.82 & 83.55 & 73.18 \\
Ray-SOTA (VFF) & 88.23 & 86.34 & 85.98 & 88.13 & 87.95 & 87.88 & 86.27 & 88.02 & 87.96 & 87.32 \\
BEV-SOTA (3D-CVF) & 87.60 & 86.17 & 83.43 & 82.46 & 82.06 & 80.55 & 80.67 & 86.30 & 85.63 & 83.41 \\
\midrule 
Virtual-fair  & 89.10 & 87.04 & 86.76 & \textbf{89.22} & 87.02 & 87.07 & 86.76 & 87.21 & 86.87 & 87.23 \\
Voxel-fair  & \textbf{91.10} & 65.25 & 62.96 & 82.01 & 67.62 & 74.70 & 60.56 & 88.82 & 83.55 & 73.18 \\
Ray-fair  & 89.16 & \textbf{88.96} & \textbf{88.96} & 89.13 & \textbf{89.02} & \textbf{89.08} & 87.05 & \textbf{89.13} & \textbf{89.04} & \textbf{88.80}\\
BEV-fair  & 88.79 & 88.66 & 88.65 & 88.79 & 88.79 & 88.76 & \textbf{88.66} & 87.66 & 86.43 & 88.30\\
\bottomrule 
\end{tabular}}
\label{tab:KITTI-C_single-modal_evaluation_BEV_AP_hard}
\end{table*}

\begin{table*}[tb]
\centering
\caption{The BEV AP results of object detectors with enhancement on car category of KITTI-C at medium difficulty.}
\resizebox{1\linewidth}{!}{
\begin{tabular}{l|c|cccccccc|c}
\toprule 
\multicolumn{1}{l|}{model} & \multicolumn{1}{c|}{clean} & \multirow{1}{*}{fog$_{low}$} & fog$_{high}$ & rain$_{low}$ & rain$_{high}$ & snow$_{low}$ & snow$_{high}$ & sunlight$_{low}$ & sunlight$_{high}$ & average\tabularnewline
\midrule 
Virtual-SOTA (VirConv) & 91.70 & 88.29 & 86.27 & \textbf{70.86} & 64.87 & 67.89 & 61.75 & \textcolor{blue}{91.62} & 85.24 & 77.11\tabularnewline
Virtual-Sigmoid & 91.70 & \textbf{88.57} & \textbf{86.73} & \textcolor{blue}{70.43} & \textcolor{blue}{67.22} & \textbf{69.83} & \textcolor{blue}{63.61} & \textbf{91.76} & \textbf{91.53} & \textbf{78.71}\tabularnewline
Virtual-Attention & 91.04 & \textcolor{blue}{88.57} & \textcolor{blue}{86.34} & 69.86 & \textbf{67.86} & \textcolor{blue}{69.65} & \textbf{67.73} & 91.03 & \textcolor{blue}{85.90} & \textcolor{blue}{78.37}\tabularnewline
\midrule 
Voxel-SOTA (FocalsConv) & 91.32 & 66.19 & 52.57 & 55.48 & 42.22 & 42.03 & 30.88 & 88.98 & 82.41 & 51.52\tabularnewline
Voxel-Sigmoid & 91.39 & \textcolor{blue}{68.60} & \textcolor{blue}{59.99} & \textbf{60.16} & \textbf{46.85} & \textbf{53.24} & \textbf{42.98} & \textcolor{blue}{89.05} & \textcolor{blue}{82.55} & \textbf{62.93} \tabularnewline
Voxel-Attention & 91.46 & \textbf{71.36} & \textbf{63.83} & \textcolor{blue}{58.49} & \textcolor{blue}{44.84} & \textcolor{blue}{45.17} & \textcolor{blue}{40.70} & \textbf{89.11} & \textbf{82.96} & \textcolor{blue}{62.06} \tabularnewline
\midrule
Ray-SOTA (VFF) & 88.23 & \textcolor{blue}{85.36} & 82.81 & \textcolor{blue}{66.71} & 62.49 & 68.81 & 59.33 & \textbf{87.97} & 73.31 & 73.25\tabularnewline
Ray-Sigmoid & 88.38 & 85.23 & \textbf{84.96} & \textbf{68.84} & \textcolor{blue}{65.27} & \textcolor{blue}{69.28} & \textcolor{blue}{60.81} & \textcolor{blue}{87.75} & \textcolor{blue}{76.63} & \textcolor{blue}{74.85}\tabularnewline
Ray-Attention & 87.91 & \textbf{85.46} & \textcolor{blue}{83.11} & 66.42 & \textbf{65.99} & \textbf{69.37} & \textbf{63.81} & 87.75 & \textbf{81.62} & \textbf{75.44}\tabularnewline
\midrule
BEV-SOTA (3D-CVF) & 88.20 & 82.27 & 77.82 & 61.53 & 56.56 & 60.85 & 54.05 & 81.51 & 76.52 & 60.73\tabularnewline
BEV-Sigmoid & 90.51 & \textcolor{blue}{87.15} & \textcolor{blue}{85.90} & \textcolor{blue}{66.09} & \textcolor{blue}{65.86} & \textcolor{blue}{65.51} & \textbf{66.58} & \textcolor{blue}{89.86} & \textbf{85.06} & \textbf{76.50}\tabularnewline
BEV-Attention & 90.57 & \textbf{87.31} & \textbf{85.91} & \textbf{66.98} & \textbf{66.53} & \textbf{67.28} & \textcolor{blue}{66.29} & \textbf{90.27} & \textcolor{blue}{78.01} & \textcolor{blue}{76.07}\tabularnewline
\bottomrule
\end{tabular}}
\label{tab:KITTI-C_sigmoid_attention_mid_BEV}
\end{table*}

\begin{table*}[tb]
\centering
\caption{The 3D AP results of object detectors with enhancement on car category of KITTI-C at easy difficulty.}
\resizebox{1\linewidth}{!}{
\begin{tabular}{l|c|cccccccc|c}
\toprule 
\multicolumn{1}{l|}{model} & \multicolumn{1}{c|}{clean} & \multirow{1}{*}{fog$_{low}$} & fog$_{high}$ & rain$_{low}$ & rain$_{high}$ & snow$_{low}$ & snow$_{high}$ & sunlight$_{low}$ & sunlight$_{high}$ & average\tabularnewline
\midrule 
Virtual-SOTA (VirConv) & 92.77 & \textbf{91.10} & \textbf{90.79} & 82.00 & 79.50 & 79.58 & 77.43 & 92.20 & 81.37 & 84.25\tabularnewline
Virtual-Sigmoid & 92.77 & 90.67 & 90.31 & \textcolor{blue}{82.07} & \textcolor{blue}{81.45} & \textbf{81.91} & \textcolor{blue}{79.27} & \textbf{92.49} & \textbf{91.15} & \textbf{86.17}\tabularnewline
Virtual-Attention & 92.74 & \textcolor{blue}{90.89} & \textcolor{blue}{90.54} & \textbf{82.37} & \textbf{81.57} & \textcolor{blue}{81.43} & \textbf{81.19} & \textcolor{blue}{92.22} & \textcolor{blue}{82.82} & \textcolor{blue}{85.38}\tabularnewline
\midrule 
Voxel-SOTA (FocalsConv) & 92.47 & \textcolor{blue}{80.24} & 72.63 & 63.72 & 57.20 & 55.08 & 41.51 & \textcolor{blue}{92.17} & 86.99 & 69.07\tabularnewline
Voxel-Sigmoid & 92.43 & \textbf{80.25} & \textcolor{blue}{74.05} & \textbf{68.11} & \textbf{59.07} & \textbf{63.33} & \textbf{57.53} & \textbf{92.30} & \textcolor{blue}{87.48} & \textbf{72.77} \tabularnewline
Voxel-Attention & 92.53 & 78.68 & \textbf{75.70} & \textcolor{blue}{67.68} & \textcolor{blue}{58.85} & \textcolor{blue}{57.34} & \textcolor{blue}{52.20} & 92.10 & \textbf{87.78} & \textcolor{blue}{71.29} \tabularnewline
\midrule
Ray-SOTA (VFF) & 89.62 & \textcolor{blue}{86.71} & \textcolor{blue}{85.73} & 74.08 & 69.72 & 75.87 & 66.64 & \textcolor{blue}{88.32} & 63.81 & 76.36\tabularnewline
Ray-Sigmoid & 89.87 & 85.25 & 80.25 & \textbf{75.75} & \textcolor{blue}{71.88} & \textcolor{blue}{76.39} & \textcolor{blue}{68.04} & \textbf{88.75} & \textcolor{blue}{66.07} & \textcolor{blue}{76.55}\tabularnewline
Ray-Attention & 89.13 & \textbf{88.03} & \textbf{87.59} & \textcolor{blue}{75.02} & \textbf{74.66} & \textbf{76.52} & \textbf{75.56} & 87.75 & \textbf{76.52} & \textbf{80.21}\tabularnewline
\midrule
BEV-SOTA (3D-CVF) & 89.17 & 84.01 & 82.79 & \textcolor{blue}{74.98} & 71.58 & \textbf{76.05} & 68.88 & 84.28 & 68.91 & 77.57\tabularnewline
BEV-Sigmoid & 91.87 & \textcolor{blue}{89.08} & \textbf{88.39} & {72.72} & \textcolor{blue}{72.37} & 73.31 & \textcolor{blue}{72.53} & \textcolor{blue}{90.54} & \textbf{78.20} & \textcolor{blue}{79.64}\tabularnewline
BEV-Attention & 91.90 & \textbf{89.17} & \textcolor{blue}{88.01} & \textbf{75.03} & \textbf{74.88} & \textcolor{blue}{75.55} & \textbf{74.51} & \textbf{90.83} & \textcolor{blue}{69.30} & \textbf{79.66}\tabularnewline
\bottomrule
\end{tabular}}
\label{tab:KITTI-C_sigmoid_attention_3D_easy}
\end{table*}

\begin{table*}[tb]
\centering
\caption{The BEV AP results of object detectors with enhancement on car category of KITTI-C at easy difficulty.}
\resizebox{1\linewidth}{!}{
\begin{tabular}{l|c|cccccccc|c}
\toprule 
\multicolumn{1}{l|}{model} & \multicolumn{1}{c|}{clean} & \multirow{1}{*}{fog$_{low}$} & fog$_{high}$ & rain$_{low}$ & rain$_{high}$ & snow$_{low}$ & snow$_{high}$ & sunlight$_{low}$ & sunlight$_{high}$ & average\tabularnewline
\midrule 
Virtual-SOTA (VirConv) & 95.96 & 94.29 & 92.93 & \textcolor{blue}{87.01} & 85.06 & 86.21 & 84.06 & \textcolor{blue}{95.37} & 85.67 & 88.82\tabularnewline
Virtual-Sigmoid & 95.96 & \textbf{95.18} & \textbf{94.71} & {86.40} & \textcolor{blue}{86.14} & \textbf{87.92} & \textcolor{blue}{85.65} & \textbf{95.58} & \textbf{94.58} & \textbf{90.77}\tabularnewline
Virtual-Attention & 95.61 & \textcolor{blue}{94.80} & \textcolor{blue}{93.73} & \textbf{87.85} & \textbf{87.82} & \textcolor{blue}{87.90} & \textbf{88.10} & 95.16 & \textcolor{blue}{86.81} & \textcolor{blue}{90.27}\tabularnewline
\midrule 
Voxel-SOTA (FocalsConv) & 93.22 & 87.90 & 80.03 & 73.40 & 64.81 & 62.38 & 46.77 & \textcolor{blue}{93.15} & 91.07 & 74.94\tabularnewline
Voxel-Sigmoid & 93.39 & \textcolor{blue}{88.05} & \textcolor{blue}{83.31} & \textcolor{blue}{76.92} & \textbf{69.43} & \textbf{73.56} & \textbf{65.94} & \textbf{93.30} & 90.98 & \textbf{80.19} \tabularnewline
Voxel-Attention & 95.33 & \textbf{88.83} & \textbf{85.57} & \textbf{78.11} & \textcolor{blue}{68.10} & \textcolor{blue}{66.13} & \textcolor{blue}{61.78} & 93.14 & \textcolor{blue}{91.01} & \textcolor{blue}{79.08} \tabularnewline
\midrule
Ray-SOTA (VFF) & 92.77 & \textbf{91.38} & \textcolor{blue}{90.34} & 83.02 & 79.89 & {84.38} & 77.66 & \textbf{91.61} & 70.74 & 83.63\tabularnewline
Ray-Sigmoid & 92.56 & 89.30 & {87.20} & \textbf{84.81} & \textbf{82.34} & \textbf{85.20} & \textcolor{blue}{78.09} & \textcolor{blue}{91.28} & \textcolor{blue}{73.50} & \textcolor{blue}{83.97}\tabularnewline
Ray-Attention & 92.38 & \textcolor{blue}{91.30} & \textbf{90.64} & 83.61 & \textcolor{blue}{82.26} & \textcolor{blue}{84.46} & \textbf{82.99} & 91.28 & \textbf{83.40} & \textbf{86.24}\tabularnewline
\midrule
BEV-SOTA (3D-CVF) & 90.41 & 90.12 & 88.18 & \textbf{82.70} & 79.22 & 83.68 & 78.20 & 85.44 & 74.67 & 82.78\tabularnewline
BEV-Sigmoid & 95.03 & \textbf{92.45} & \textcolor{blue}{92.07} & {82.20} & \textcolor{blue}{82.14} & \textcolor{blue}{84.24} & \textcolor{blue}{81.69} & \textcolor{blue}{93.66} & \textbf{81.85} & \textbf{86.29}\tabularnewline
BEV-Attention & 95.13 & \textcolor{blue}{92.22} & \textbf{91.38} & \textcolor{blue}{82.53} & \textbf{82.30} & \textbf{84.97} & \textbf{81.98} & \textbf{94.31} & \textcolor{blue}{76.15} & \textcolor{blue}{85.73}\tabularnewline
\bottomrule
\end{tabular}}
\label{tab:KITTI-C_sigmoid_attention_easy_BEV}
\end{table*}

\begin{table*}[tb]
\centering
\caption{The 3D AP results of object detectors with enhancement on car category of KITTI-C at hard difficulty.}
\resizebox{1\linewidth}{!}{
\begin{tabular}{l|c|cccccccc|c}
\toprule 
\multicolumn{1}{l|}{model} & \multicolumn{1}{c|}{clean} & \multirow{1}{*}{fog$_{low}$} & fog$_{high}$ & rain$_{low}$ & rain$_{high}$ & snow$_{low}$ & snow$_{high}$ & sunlight$_{low}$ & sunlight$_{high}$ & average\tabularnewline
\midrule 
Virtual-SOTA (VirConv) & 85.60 & \textbf{76.77} & \textcolor{blue}{72.99} & 55.55 & 52.81 & \textcolor{blue}{54.84} & 50.84 & \textcolor{blue}{84.96} & 74.10 & 65.36\tabularnewline
Virtual-Sigmoid & 85.60 & \textcolor{blue}{76.58} & {72.50} & \textbf{59.60} & \textcolor{blue}{57.27} & 53.33 & \textbf{63.61} & \textbf{91.76} & \textbf{80.28} & \textbf{69.37}\tabularnewline
Virtual-Attention & 85.03 & {76.03} & \textbf{73.24} & \textcolor{blue}{59.07} & \textbf{57.48} & \textbf{57.30} & \textcolor{blue}{55.32} & 84.33 & \textcolor{blue}{74.93} & \textcolor{blue}{67.21}\tabularnewline
\midrule 
Voxel-SOTA (FocalsConv) & 85.07 & \textcolor{blue}{53.25} & 44.09 & 42.68 & 33.94 & 33.89 & 25.11 & \textbf{82.97} & 74.09 & 48.75\tabularnewline
Voxel-Sigmoid & 83.41 & \textbf{54.86} & \textcolor{blue}{47.56} & \textbf{46.75} & \textbf{36.09} & \textbf{42.12} & \textbf{32.39} & \textcolor{blue}{82.61} & \textcolor{blue}{74.52} & \textbf{52.11} \tabularnewline
Voxel-Attention & 83.39 & {52.74} & \textbf{50.13} & \textcolor{blue}{43.18} & \textcolor{blue}{34.55} & \textcolor{blue}{34.76} & \textcolor{blue}{30.21} & 82.61 & \textbf{74.67} & \textcolor{blue}{50.36} \tabularnewline
\midrule
Ray-SOTA (VFF) & 80.18 & 69.60 & \textcolor{blue}{66.78} & 50.72 & 44.37 & \textcolor{blue}{51.46} & 42.08 & \textbf{79.57} & 60.59 & 58.15\tabularnewline
Ray-Sigmoid & 81.69 & \textcolor{blue}{69.70} & {60.19} & \textcolor{blue}{51.41} & \textcolor{blue}{46.64} & {50.64} & \textcolor{blue}{44.34} & \textcolor{blue}{79.21} & \textcolor{blue}{63.96} & \textcolor{blue}{58.26}\tabularnewline
Ray-Attention & 79.58 & \textbf{73.77} & \textbf{72.24} & \textbf{52.03} & \textbf{51.48} & \textbf{52.32} & \textbf{51.12} & 79.21 & \textbf{69.06} & \textbf{62.65}\tabularnewline
\midrule
BEV-SOTA (3D-CVF) & 78.12 & 67.44 & 63.53 & 46.14 & 43.07 & 45.79 & 41.23 & 71.70 & 65.36 & 55.53\tabularnewline
BEV-Sigmoid & 82.25 & \textcolor{blue}{76.21} & \textcolor{blue}{73.84} & \textcolor{blue}{51.29} & \textcolor{blue}{51.22} & \textcolor{blue}{51.69} & \textcolor{blue}{51.09} & \textcolor{blue}{79.77} & \textbf{75.87} & \textcolor{blue}{63.87}\tabularnewline
BEV-Attention & 82.14 & \textbf{76.68} & \textbf{75.38} & \textbf{53.59} & \textbf{53.36} & \textbf{52.91} & \textbf{52.35} & \textbf{82.01} & \textcolor{blue}{67.07} & \textbf{64.17}\tabularnewline
\bottomrule
\end{tabular}}
\label{tab:KITTI-C_sigmoid_attention_3D_hard}
\end{table*}

\begin{table*}[tb]
\centering
\caption{The BEV AP results of object detectors with enhancement on car category of KITTI-C at hard difficulty.}
\resizebox{1\linewidth}{!}{
\begin{tabular}{l|c|cccccccc|c}
\toprule 
\multicolumn{1}{l|}{model} & \multicolumn{1}{c|}{clean} & \multirow{1}{*}{fog$_{low}$} & fog$_{high}$ & rain$_{low}$ & rain$_{high}$ & snow$_{low}$ & snow$_{high}$ & sunlight$_{low}$ & sunlight$_{high}$ & average\tabularnewline
\midrule 
Virtual-SOTA (VirConv) & 91.33 & 83.53 & \textcolor{blue}{81.58} & 64.27 & 60.72 & 63.47 & 57.16 & {89.26} & 80.78 & 72.60\tabularnewline
Virtual-Sigmoid & 91.33 & \textbf{85.69} & \textbf{81.61} & \textbf{68.31} & \textcolor{blue}{63.84} & \textbf{66.04} & \textcolor{blue}{59.56} & \textcolor{blue}{89.58} & \textbf{88.00} & \textbf{75.33}\tabularnewline
Virtual-Attention & 91.03 & \textcolor{blue}{83.73} & {80.90} & \textcolor{blue}{67.70} & \textbf{65.66} & \textcolor{blue}{65.78} & \textbf{63.80} & \textbf{90.63} & \textcolor{blue}{83.48} & \textcolor{blue}{75.21}\tabularnewline
\midrule 
Voxel-SOTA (FocalsConv) & 91.10 & 61.89 & 48.06 & 51.39 & 37.92 & 37.76 & 28.67 & {88.34} & 80.53 & 54.32\tabularnewline
Voxel-Sigmoid & 91.11 & \textcolor{blue}{64.22} & \textcolor{blue}{55.88} & \textbf{56.12} & \textbf{42.49} & \textbf{49.17} & \textbf{38.39} & \textbf{88.89} & \textcolor{blue}{80.67} & \textbf{59.48} \tabularnewline
Voxel-Attention & 89.36 & \textbf{64.60} & \textbf{59.29} & \textcolor{blue}{52.25} & \textcolor{blue}{40.31} & \textcolor{blue}{41.05} & \textcolor{blue}{36.18} & \textcolor{blue}{88.86} & \textbf{81.03} & \textcolor{blue}{57.95} \tabularnewline
\midrule
Ray-SOTA (VFF) & 88.23 & 80.22 & \textcolor{blue}{77.62} & 61.88 & 55.70 & \textcolor{blue}{62.71} & 53.41 & \textbf{87.69} & 71.67 & 68.83\tabularnewline
Ray-Sigmoid & 88.15 & \textcolor{blue}{80.67} & {68.00} & \textcolor{blue}{62.82} & \textcolor{blue}{58.40} & 62.40 & \textcolor{blue}{55.74} & \textcolor{blue}{87.43} & \textcolor{blue}{75.51} & \textcolor{blue}{68.87}\tabularnewline
Ray-Attention & 87.65 & \textbf{82.28} & \textbf{79.47} & \textbf{63.16} & \textbf{62.95} & \textbf{64.22} & \textbf{60.95} & 87.43 & \textbf{80.07} & \textbf{72.57}\tabularnewline
\midrule
BEV-SOTA (3D-CVF) & 87.60 & 77.05 & 72.83 & 54.75 & 49.88 & 54.09 & 47.39 & 78.32 & 69.31 & 62.95\tabularnewline
BEV-Sigmoid & 88.63 & \textcolor{blue}{84.60} & \textcolor{blue}{82.20} & \textcolor{blue}{62.43} & \textcolor{blue}{62.29} & \textcolor{blue}{63.00} & \textcolor{blue}{62.18} & \textcolor{blue}{88.17} & \textbf{85.64} & \textcolor{blue}{73.81}\tabularnewline
BEV-Attention & 89.02 & \textbf{84.94} & \textbf{83.46} & \textbf{64.83} & \textbf{64.22} & \textbf{64.02} & \textbf{63.46} & \textbf{88.44} & \textcolor{blue}{78.07} & \textbf{73.93}\tabularnewline
\bottomrule
\end{tabular}}
\label{tab:KITTI-C_sigmoid_attention_hard_BEV}
\end{table*}

\section{Visualization of Comparison between SOTA Fusion Models and Our Enhancement Methods}\label{sec:visualization_entire}
\begin{figure*}[t]
\centering
\includegraphics[width=1\linewidth]{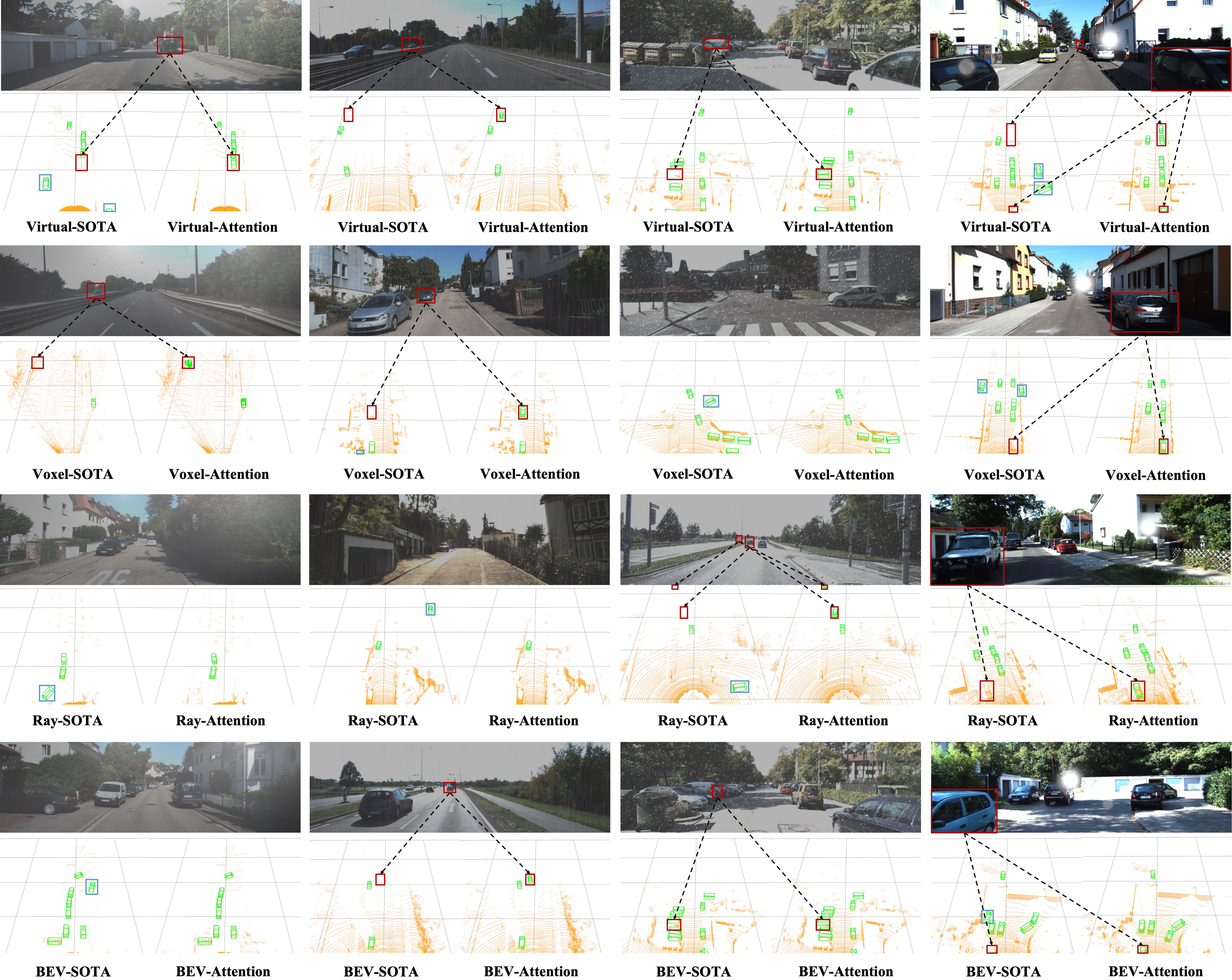}
\caption{Visualization comparison between SOTA fusion models and corresponding enhanced ones. Red boxes represent missed detections and blue boxes represent misdetections.}
\label{fig:Visualization_of_results_all_Attention}
\end{figure*}

In Figure~\ref{fig:Visualization_of_results_all_Attention}, we visualize more results of the comparison between SOTA models and attention-based enhanced models. The objects in red boxes are undetected by SOTA models while objects in blue boxes are false detections of SOTA models.

\begin{figure*}[t]
\centering
\includegraphics[width=1\linewidth]{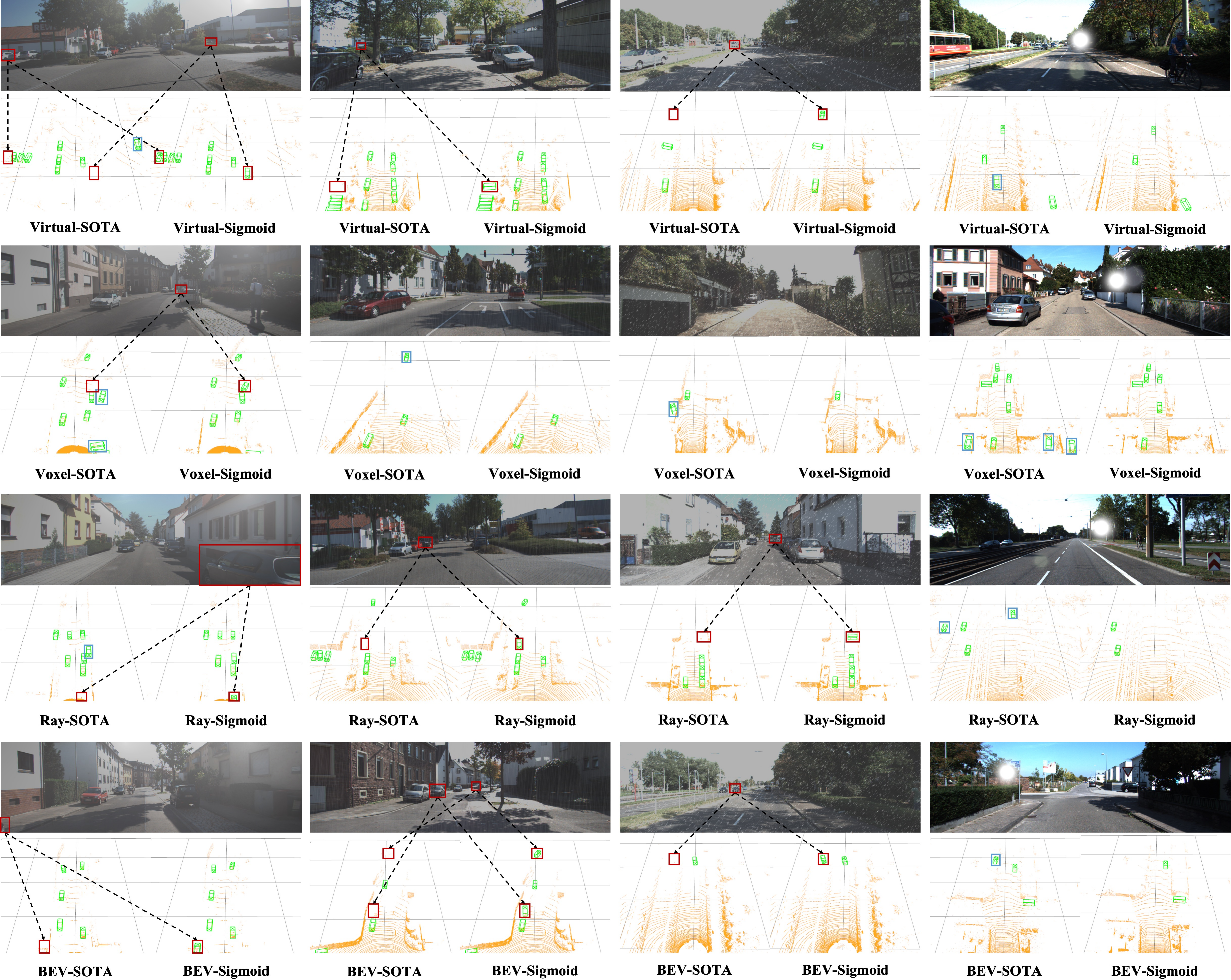}
\caption{Visualization comparison between SOTA fusion models and corresponding enhanced ones. Red boxes represent missed detections and blue boxes represent misdetections.}
\label{fig:Visualization_of_results_all_sigmoid}
\end{figure*}

In Figure~\ref{fig:Visualization_of_results_all_sigmoid}, we visualize more results of the comparison between SOTA models and Sigmoid-based enhanced models. The objects in red boxes are undetected by SOTA models while objects in blue boxes are false detections of SOTA models.

\section{Conclusion}
We analyze the robustness from the perspective of the fusion modals and find virtual point-based fusion model is the best one. Also, we find that flexibly weighted fusing the features from LiDAR and camera branches can effectively improve the robustness. In future work, we aim to analyze why the virtual point-based model is the best and try to supplement its advantage with other modals to make them better.

\section{Impact Statement}
This paper presents work whose goal is to advance the field of Machine Learning. There are many potential societal consequences of our work, none of which we feel must be specifically highlighted here.

%
%
%
%


%

\ifCLASSOPTIONcaptionsoff
  \newpage
\fi



%

\bibliographystyle{IEEEtran}
\bibliography{ref}

\end{document}